\documentclass{article}
\usepackage[preprint]{neurips_2026}

\usepackage[utf8]{inputenc}
\usepackage[T1]{fontenc}
\usepackage{hyperref}
\usepackage{url}
\usepackage{booktabs}
\usepackage{amsfonts}
\usepackage{amsmath,amssymb,amsthm}
\newtheorem{theorem}{Theorem}

\newtheorem{proposition}[theorem]{Proposition}

\newtheorem{definition}[theorem]{Definition}
\usepackage{algorithm}
\usepackage{algorithmic}
\usepackage{multirow}
\usepackage{float}
\usepackage{nicefrac}
\usepackage{microtype}
\usepackage{xcolor}
\usepackage{graphicx}

\title{Quotient DAGs for Off-Policy Evaluation:\\Forward-Flow Importance Sampling and Exact Slate Propensities}

\author{
  Ziwen Xie\thanks{Equal contribution.} \\
  Shanghai Jiao Tong University \\
  \texttt{shuji888@sjtu.edu.cn}
  \And
  Shaowen Xiang\footnotemark[1] \\
  Shanghai Jiao Tong University \\
  \texttt{starryspace@sjtu.edu.cn}
  \And
  Hongyu He \\
  National University of Singapore \\
  \texttt{hongyu.h@nus.edu.sg}
  \And
  Dianbo Liu\thanks{Corresponding author.} \\
  National University of Singapore \\
  \texttt{dianbo@nus.edu.sg}
}

\begin{document}
\maketitle

\begin{abstract}
Off-policy evaluation estimates how a target policy would perform
using data collected by a different behavior policy, which is
crucial when online testing is costly or risky, such as in
recommendation or healthcare. Standard importance sampling
reweights each logged trajectory, but it can treat details of the
generation process as meaningful even when the evaluation target
ignores them: for example, an autoregressive slate recommender may
generate an ordered sequence of items while the reward and
downstream estimator depend only on the unordered slate. This
creates nuisance variance and a computational gap, since exact
unordered slate propensities require summing over all generation
orders. We introduce a quotient-DAG view that merges histories
equivalent for evaluation and assigns weights using
target-to-behavior forward-flow ratios on the merged graph. For
slate recommendation under a set-sufficient next-item interface,
this yields Forward-DP, a subset-DAG dynamic program that computes
exact unordered propensities without factorial enumeration. The
resulting propensity primitive enables practical propensity-based
evaluation and model selection for context-dependent autoregressive
slate loggers.
\end{abstract}

\section{Introduction}
\label{sec:intro}

Reinforcement learning is widely used for sequential decision policies
in recommendation, clinical treatment, and public-sector resource
allocation. In many such settings, deploying a new policy directly is
expensive or unsafe: a candidate sepsis-treatment protocol, for
instance, cannot be tested on ICU patients merely to see whether it
improves outcomes. Off-policy evaluation (OPE) addresses this problem
by estimating the value of a target policy from data logged under a
different behavior policy. Importance sampling (IS), the canonical
unbiased OPE estimator, reweights each logged trajectory by a product of
target-to-behavior action-probability ratios, but these products can
vary by orders of magnitude as horizon and policy mismatch grow.

\begin{figure*}[t]
    \centering
    \includegraphics[width=0.9\textwidth]{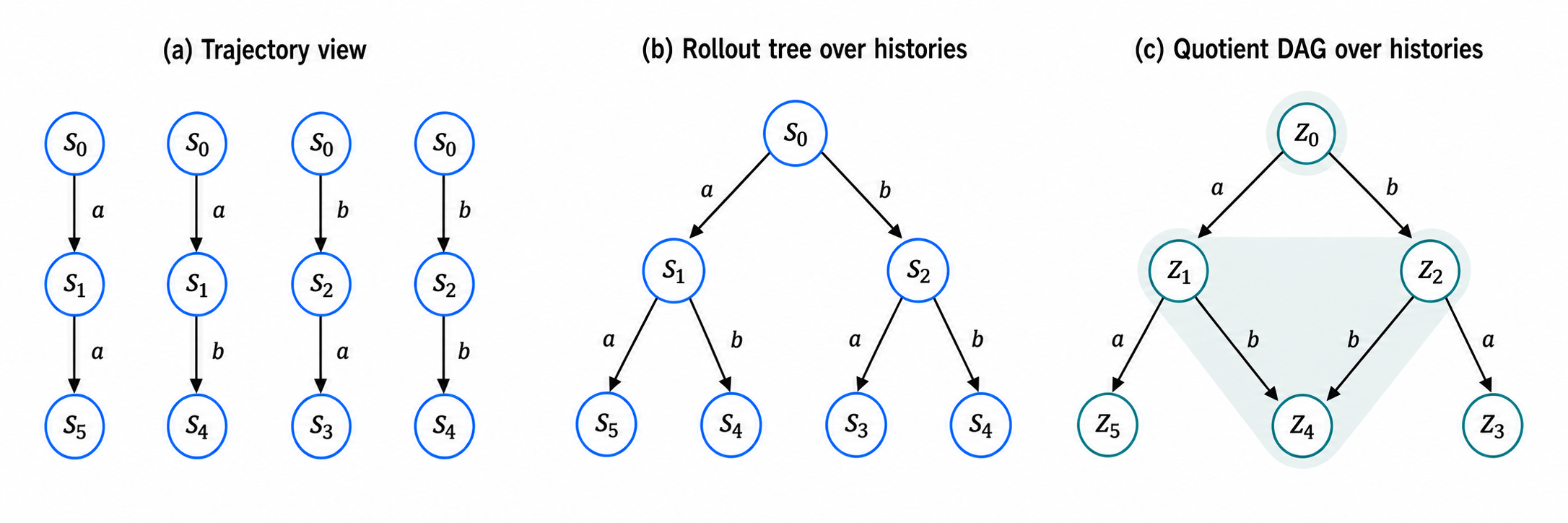}
    \vspace{-0.5em}
    \caption{
    \textbf{From trajectory IS to quotient-DAG weighting.}
    \textbf{(a)} Trajectory IS assigns a likelihood ratio to each realized
    path. \textbf{(b)} The rollout tree shares prefixes but still represents
    distinct history-prefix nodes, even with repeated state labels.
    \textbf{(c)} A quotient DAG merges evaluation-equivalent prefixes into
    multi-parent nodes such as \(z_4\). The quotient likelihood ratio is the
    forward-flow ratio \(F_\pi(z)/F_\beta(z)\), not any single path ratio.
    }
    \label{fig:quotient-dag-overview}
    \vspace{-0.8em}
\end{figure*}

This work starts from the observation, illustrated in
Figure~\ref{fig:quotient-dag-overview}, that the rollout tree can be a
finer representation than the evaluation problem requires. A logged
trajectory is one path through the tree, and trajectory IS assigns a
likelihood ratio to that particular path. Yet the tree may contain
history prefixes that are equivalent for the relevant continuation or
evaluation target. We quotient the rollout tree by such an equivalence
relation, collapsing equivalent prefixes into nodes of a layered
directed acyclic graph (DAG). On this quotient DAG, the probability
mass reaching a node under each policy defines its forward flow, and
the appropriate likelihood ratio is the target-to-behavior flow ratio
$F_\pi(z)/F_\beta(z)$. Per-decision IS, marginalized IS, and
known-abstraction MIS are special cases of this construction,
corresponding to different choices of equivalence relation.

Our main concrete instantiation is slate recommendation, where modern
slate generators often construct slates autoregressively, exposing
per-step probabilities along one realized ordering. Yet slate rewards
and propensity-consuming slate-OPE estimators may depend only on the
unordered slate. This creates a joint-propensity gap: the logger
exposes an ordering probability, while the estimator requires the
total probability of the unordered slate. Computing this quantity
exactly requires summing over all generation orders, and fixed-score
Plackett--Luce expressions no longer apply once the next-item softmax
depends on the partial slate.

In this paper, we formulate the quotient-space likelihood ratio as a
forward-flow ratio and use it to construct forward-flow IS estimators.
For terminal quotient-measurable returns, this gives an exact
variance-gap expression that identifies the within-class
likelihood-ratio variation removed by quotient weighting. For
autoregressive slate generation, when the per-step distribution depends
only on the context and the set of items already picked, forward flow
on the subset DAG computes the exact unordered slate propensity in
time polynomial in the catalog size and exponential in the slate size.
This closes the joint-propensity computational gap for
context-dependent set-sufficient loggers. We evaluate the resulting
estimators on finite-horizon MDP benchmarks and use KuaiRec slate
experiments to test the Forward-DP propensity primitive in downstream
OPE and model selection.

\section{Problem setup}
\label{sec:setup}

\paragraph{Off-policy evaluation.}
We consider off-policy evaluation (OPE) of a target policy $\pi$ from
data logged under a behavior policy $\beta$. The environment is a
finite-horizon MDP
$\mathcal{M}=(\mathcal{S},\mathcal{A},P,R,d_0,\gamma,H)$.
A trajectory
$\tau=(s_1,a_1,r_1,\ldots,s_H,a_H,r_H)$ is generated by
$s_1\sim d_0$, $a_t\sim\pi(\cdot\mid s_t)$,
$r_t\sim R(s_t,a_t)$, and $s_{t+1}\sim P(\cdot\mid s_t,a_t)$
for $t<H$. The discounted return is
$J(\tau)=\sum_{t=1}^{H}\gamma^{t-1}r_t$, and the policy value is
$V(\pi)=\mathbb{E}_\pi[J(\tau)]$.

Given logged trajectories
$\mathcal{D}=\{\tau^{(i)}\}_{i=1}^{N}\sim\beta$, define the
per-step ratio
$\rho_t(\tau)=\pi(a_t\mid s_t)/\beta(a_t\mid s_t)$ and
$\rho_{1:t}(\tau)=\prod_{t'=1}^{t}\rho_{t'}(\tau)$. The
trajectory-level importance sampling estimator is
\begin{equation}
\widehat V_{\mathrm{IS}}(\pi)
=
\frac{1}{N}\sum_{i=1}^{N}
\rho_{1:H}(\tau^{(i)})J(\tau^{(i)}),
\label{eq:traj-is}
\end{equation}
which is unbiased under the standard support condition that the target
trajectory distribution is absolutely continuous with respect to the
behavior trajectory distribution~\citep{precup2000eligibility}. The
self-normalized variant
$\widehat V_{\mathrm{WIS}}(\pi)
=\sum_i w_i J(\tau^{(i)}) / \sum_i w_i$ with
$w_i=\rho_{1:H}(\tau^{(i)})$ trades finite-sample unbiasedness for
typically lower variance~\citep{mahmood2014weighted}.

\paragraph{Slate OPE.}
The slate setting is a contextual combinatorial bandit at the
evaluation level. A context $x\sim p_{\mathcal X}$ is observed, a
candidate pool $[M]=\{1,\ldots,M\}$ is available, and the policy
selects a size-$K$ slate $S\subseteq[M]$ with $|S|=K$. We assume the
evaluation reward is a set function $r(x,S)$, invariant to the
generation order of the slate. The value of a slate policy is
\[
V(\pi)=
\mathbb{E}_{x\sim p_{\mathcal X},\,S\sim\pi(\cdot\mid x)}
[r(x,S)].
\]

Although evaluation is slate-level, the logger may generate $S$
autoregressively as an ordered sequence
$\sigma=(\sigma_1,\ldots,\sigma_K)$ with
$S=\{\sigma_1,\ldots,\sigma_K\}$. Let
$h_{<t}=(\sigma_1,\ldots,\sigma_{t-1})$ be the ordered prefix and
$S_{<t}=\{\sigma_1,\ldots,\sigma_{t-1}\}$ the picked set. For a
slate policy $\mu\in\{\beta,\pi\}$,
\begin{equation}
\mu(\sigma\mid x)=
\prod_{t=1}^{K}\mu(\sigma_t\mid x,h_{<t}).
\label{eq:general-order-prop}
\end{equation}

\begin{definition}[Set-sufficient slate policy]
\label{def:set-sufficiency}
A slate policy $\mu$ is \emph{set-sufficient} if, for every context
$x$, step $t$, item $a$, and any two ordered prefixes
$h_{<t}$ and $h'_{<t}$ with the same picked set $S_{<t}$,
\begin{equation}
\mu(a\mid x,h_{<t})=\mu(a\mid x,h'_{<t})
\qquad
\text{for all } a\notin S_{<t}.
\label{eq:set-sufficiency}
\end{equation}
We write this common value as $\mu(a\mid x,S_{<t})$.
\end{definition}

Our exact subset-DAG construction assumes that the behavior policy
$\beta$ and every target policy under evaluation are set-sufficient.
Under set-sufficiency, an autoregressive Plackett--Luce policy can be
written as
\begin{equation}
\mu_\theta(a\mid x,S_{<t})
=
\frac{
\exp\!\left(\ell_\theta(a;\mathrm{prompt}(x,S_{<t}))/T_{\mathrm{temp}}\right)
}{
\sum_{a'\notin S_{<t}}
\exp\!\left(\ell_\theta(a';\mathrm{prompt}(x,S_{<t}))/T_{\mathrm{temp}}\right)
},
\label{eq:transformer-pl}
\end{equation}
where $\mathrm{prompt}(x,S_{<t})$ is the model's input encoding of the
context $x$ together with the picked set $S_{<t}$, $\ell_\theta(a;\cdot)$
is the model's logit for item $a$, and $T_{\mathrm{temp}}>0$ is the
sampling temperature.

\paragraph{Scope of set-sufficiency.}
Set-sufficiency is imposed on the slate-generation interface, not on
the architecture alone. It holds for score-then-mask deployments in
which a model scores the remaining candidates and previously selected
items affect future choices only through the exclusion mask; this is
the canonical slate-generation lift adopted in practice for SASRec,
BERT4Rec, and NeuralPGRank-style neural rankers
\citep{kang2018sasrec, sun2019bert4rec, gao2023neuralpgrank}.
For prompt-based LLM rerankers, we enforce the same interface by
presenting the picked set in a canonical order
(\S\ref{sec:exp-transformer-ms}). Models that autoregressively
condition on the emitted order of previous items fall outside this
assumption unless their prefix representation is canonicalized, which
changes the deployed policy by removing ordering information
\citep{geng2022p5, cui2022m6rec, sun2023rankgpt}.

The induced unordered-slate propensity is
\begin{equation}
\mu(S\mid x)
=
\sum_{\sigma\in\mathrm{perm}(S)}
\mu(\sigma\mid x),
\label{eq:slate-prop}
\end{equation}
a sum over all $K!$ orderings of $S$. Slate-OPE estimators assume
access to this quantity for both
$\beta$ and $\pi$~\citep{swaminathan2017off, vlassis2021control,
kiyohara2023off, kiyohara2024slate, shimizu2024effective}.

The logged slate dataset is
$\mathcal{D}=\{(x_n,S_n,R_n)\}_{n=1}^{N}$, with
$x_n\sim p_{\mathcal X}$, $S_n\sim\beta(\cdot\mid x_n)$, and
$\mathbb{E}[R_n\mid x_n,S_n]=r(x_n,S_n)$. The slate-level IS
estimator is
\begin{equation}
\widehat V_{\mathrm{slate}}(\pi)
=
\frac{1}{N}\sum_{n=1}^{N}
\frac{\pi(S_n\mid x_n)}{\beta(S_n\mid x_n)}R_n,
\label{eq:slate-is}
\end{equation}
which is unbiased under the slate-level support condition
$\beta(S\mid x)>0$ whenever $\pi(S\mid x)>0$ for
$p_{\mathcal X}$-almost every $x$. Its self-normalized variant
estimates
$\sum_n w_nR_n/\sum_n w_n$ with
$w_n=\pi(S_n\mid x_n)/\beta(S_n\mid x_n)$, trading finite-sample
unbiasedness for typically lower variance. The doubly-robust (DR)
variant adds a control variate $\widehat r(x,S)$ to the IS residual;
with exact propensities and a fixed or independently fitted baseline,
DR is unbiased and can reduce variance when $\widehat r$ tracks the
reward~\citep{dudik2011doubly, jiang2016doubly, thomas2016data}.

\section{Method}
\label{sec:method}

We formulate IS on the rollout tree of history prefixes and then
quotient this tree by equivalence classes that are sufficient for
evaluation. Pushing both the target and behavior measures through the
quotient map gives a coarser sample space on which the correct
likelihood ratio is the ratio of forward flows,
$F_\pi(z)/F_\beta(z)$, rather than the likelihood ratio of any single
realized path. This is a graph-structured specialization of
conditional IS
\citep{rowland2020conditional, liu2020understanding}: the identity
history quotient gives per-decision IS, the state--time quotient gives
finite-horizon marginalized IS
\citep{xie2019towards, liu2020understanding}, and sufficient
abstract-state quotients recover known-abstraction MIS
\citep{pavse2023scaling}.

Our main algorithmic use of this view is autoregressive slate OPE.
When the next-item distribution depends on the context and picked set
but not on the order of the picked prefix, the permutation quotient is
the subset DAG. Forward-DP computes the exact unordered-slate
propensity of a logged slate in $O((M+K)\cdot 2^K)$ time, avoiding the
$K!$ enumeration over generation orders.

\subsection{Quotient likelihood ratios and forward-flow IS}
\label{sec:method-tree}

\paragraph{Rollout tree.}
At decision step $t$, define the pre-action history
\[
h_t=(s_1,a_1,r_1,\ldots,s_{t-1},a_{t-1},r_{t-1},s_t),
\qquad t=1,\ldots,H,
\]
with $h_1=s_1$. The rollout tree has one node for each such history;
an edge from $h_t$ to $h_{t+1}$ corresponds to drawing $a_t$,
observing $r_t$, and transitioning to $s_{t+1}$. Trajectory IS assigns
a likelihood ratio to the sampled root-to-leaf path.

\paragraph{Sufficient quotient.}
We quotient the rollout tree by an equivalence relation on pre-action
histories.

\begin{definition}[Sufficient equivalence]
\label{def:sufficient}
An equivalence relation $\sim$ on pre-action histories is
\emph{sufficient} if, for every layer $t$, every pair
$h_t\sim h'_t$, and every policy $\mu\in\{\beta,\pi\}$, the
conditional distribution under $\mu$ of the continuation
\[
(a_t,r_t,s_{t+1},a_{t+1},r_{t+1},\ldots,s_H,a_H,r_H)
\]
given $h_t$ is the same as that given $h'_t$.
\end{definition}

Let $Z_t$ be the layer-$t$ equivalence classes and
$z_t(h_t)=[h_t]$. The quotient is a layered DAG whose nodes are
classes $z\in Z_t$, with an edge $z\to z'$ whenever some history in
$z$ can transition to a history in $z'$. Sufficiency makes the
quotient-level transition probability $P_\mu(z'\mid z)$ well defined.

\paragraph{Forward flow.}
For policy $\mu\in\{\beta,\pi\}$, the forward flow into a quotient
node $z\in Z_t$ is
\begin{equation}
F_\mu(z)
=
\sum_{h_t\in z}P_\mu(h_t)
=
\mathbb{P}_\mu[z_t=z],
\label{eq:forward-flow-general}
\end{equation}
with recursion
\begin{equation}
F_\mu(z')
=
\sum_{z\in Z_t:\,z\to z'}
F_\mu(z)P_\mu(z'\mid z),
\qquad z'\in Z_{t+1}.
\label{eq:flow-recursion}
\end{equation}
Thus, given the quotient DAG and quotient transitions, all flows are
computed by a single forward sweep.

\begin{proposition}[Quotient likelihood ratio]
\label{prop:quotient-rn}
Fix a layer $t$ and assume the history-level support condition
$P_\pi^{H_t}\ll P_\beta^{H_t}$, that is, $P_\beta(h_t)=0$ implies
$P_\pi(h_t)=0$ for all pre-action histories $h_t$, so that the prefix
ratio $\rho_{1:t-1}=dP_\pi^{H_t}/dP_\beta^{H_t}$ is well-defined
$P_\beta$-a.s. Let $P_{\mu,t}^{Z}$ denote the pushforward of the
policy-induced distribution over $h_t$ under $z_t$. Then
$P_{\pi,t}^{Z}\ll P_{\beta,t}^{Z}$ and
\begin{equation}
\frac{dP_{\pi,t}^{Z}}{dP_{\beta,t}^{Z}}(z)
=
\frac{F_\pi(z)}{F_\beta(z)}.
\label{eq:quotient-rn}
\end{equation}
Equivalently, for every class $z$ with $F_\beta(z)>0$,
\begin{equation}
\mathbb{E}_\beta[
\rho_{1:t-1}(\tau)\mid z_t(\tau)=z]
=
\frac{F_\pi(z)}{F_\beta(z)} ,
\label{eq:flow-ratio}
\end{equation}
with $\rho_{1:0}\equiv 1$.
\end{proposition}

The proof is the standard pushforward likelihood-ratio identity and is
given in Appendix~\ref{app:proof-quotient-rn}. Equation
\eqref{eq:flow-ratio} is the link to conditional and marginalized IS:
different quotients recover per-decision IS, state--time MIS, and
known-abstraction MIS. The distinction here is that the conditioning
variable is represented as a history-prefix quotient with a computable
forward flow.

\paragraph{Forward-flow IS.}
For general finite-horizon OPE, we use this quotient ratio in the
per-decision estimator. Under sufficiency, the current action
distribution is class-measurable, so
\[
\rho_t(z_t,a_t)=
\frac{\pi(a_t\mid z_t)}{\beta(a_t\mid z_t)}.
\]
Forward-Flow IS replaces the sampled prefix ratio in PDIS by the
quotient likelihood ratio:
\begin{equation}
\widehat V_{\mathrm{FF}}(\pi)
=
\frac{1}{N}
\sum_{i=1}^{N}
\sum_{t=1}^{H}
\gamma^{t-1}
\frac{F_\pi(z_t^{(i)})}{F_\beta(z_t^{(i)})}
\rho_t(z_t^{(i)},a_t^{(i)})
r_t^{(i)}.
\label{eq:ff-is}
\end{equation}
Each summand replaces the history-prefix ratio by its conditional
expectation within the quotient class, while keeping the current
action correction explicit. Hence FF-IS is unbiased under the usual
support condition for PDIS, and each per-decision summand has no larger
variance than its PDIS counterpart. This is only a per-term statement;
it does not imply a full-estimator variance ordering for arbitrary
multi-step returns because cross-time covariance terms may change.

\paragraph{Terminal quotient and variance gap.}
A full Rao--Blackwell variance comparison is available when the return
is measurable with respect to a terminal quotient. Let $q$ be a
terminal quotient map, such as $q=z_H$ or, for slate OPE,
$q(\tau)=S$. Suppose $G(\tau)=g(q(\tau))$, and define
\[
w(q)=\frac{dP_\pi^q}{dP_\beta^q}(q)
=\frac{F_\pi(q)}{F_\beta(q)}.
\]
The terminal forward-flow estimator is
\begin{equation}
\widehat V_{\mathrm{FF\text{-}terminal}}(\pi)
=
\frac{1}{N}
\sum_{i=1}^{N}
w(q(\tau^{(i)}))g(q(\tau^{(i)})).
\label{eq:ff-terminal}
\end{equation}

\begin{proposition}[Exact variance gap on a quotient]
\label{prop:variance-gap}
If $G(\tau)=g(q(\tau))$ is quotient-measurable, then
\begin{align}
&\mathrm{Var}_\beta(\rho(\tau)G(\tau))
-
\mathrm{Var}_\beta(w(q(\tau))g(q(\tau)))
\nonumber\\
&\qquad =
\mathbb{E}_\beta\!\left[
g(q(\tau))^2
\mathrm{Var}_\beta(\rho(\tau)\mid q(\tau))
\right].
\label{eq:variance-gap}
\end{align}
For a finite quotient space, this equals
\begin{equation}
\sum_z
F_\beta(z)g(z)^2w(z)^2
\chi^2\!\left(
P_\pi(\cdot\mid z)\,\|\,P_\beta(\cdot\mid z)
\right).
\label{eq:chi2-gap}
\end{equation}
\end{proposition}

The proof is in Appendix~\ref{app:proof-variance-gap}. This result
identifies what terminal forward-flow weighting removes: the
within-class target--behavior mismatch. Slate OPE with
generation-order-invariant rewards is the main instance of this
terminal quotient.

\subsection{Slate instantiation: subset DAG and Forward-DP}
\label{sec:method-slate}

The slate setting of \S\ref{sec:setup} is a contextual bandit at the
evaluation level, but an autoregressive logger unfolds it into a
$K$-step tree over ordered item prefixes. Under set-sufficiency
(Definition~\ref{def:set-sufficiency}), two ordered prefixes are
equivalent whenever they contain the same picked set. The permutation
quotient is therefore the subset DAG: depth-$\ell$ nodes are subsets
$S\subseteq[M]$ with $|S|=\ell$, and edges append one unpicked item.
For a single logged slate $S_K$, only the rooted sub-DAG over subsets
of $S_K$ is needed, containing $2^K$ nodes rather than $K!$ ordered
paths.

\paragraph{Forward-DP.}
For policy $\mu$ and context $x$, define the forward flow on subsets of
a target slate by
\begin{equation}
F_\mu^x(S)
=
\begin{cases}
1, & S=\emptyset,\\[3pt]
\sum_{a\in S}
F_\mu^x(S\setminus\{a\})\mu(a\mid x,S\setminus\{a\}),
& |S|\ge 1.
\end{cases}
\label{eq:slate-forward-flow}
\end{equation}

\begin{proposition}[Forward flow equals joint slate propensity]
\label{prop:flow-exact}
If $\mu$ is set-sufficient, then
$F_\mu^x(S)=\mu(S\mid x)$ for every $S\subseteq[M]$ with
$|S|\le K$.
\end{proposition}

The proof, given in Appendix~\ref{app:proof-flow-exact}, partitions
the orderings of $S$ by their last selected item. Thus, for slate OPE,
the terminal forward-flow estimator becomes
\begin{equation}
\widehat V_{\mathrm{DAG}}(\pi)
=
\frac{1}{N}
\sum_{n=1}^{N}
\frac{F_\pi^{x_n}(S_n)}{F_\beta^{x_n}(S_n)}
R_n.
\label{eq:dag-slate-is}
\end{equation}
By Proposition~\ref{prop:flow-exact}, this is exactly the slate-level
IS estimator with the otherwise intractable joint propensities computed
by forward flow. Self-normalized and doubly robust variants replace
the same slate-level ratio by
$F_\pi^{x_n}(S_n)/F_\beta^{x_n}(S_n)$.

\paragraph{Ordering-nuisance variance gap.}
The terminal variance gap has a direct slate interpretation. Suppose
the reward distribution depends on $(x,S)$ but not on the generation
ordering, and is conditionally independent of the ordering given
$(x,S)$. Let
\[
W(x,S)=\frac{\pi(S\mid x)}{\beta(S\mid x)}
=
\frac{F_\pi^x(S)}{F_\beta^x(S)} .
\]
Then replacing trajectory IS over orderings by DAG-IS removes
\begin{equation}
\mathbb{E}_{x\sim p_X,S\sim\beta(\cdot\mid x)}
\!\left[
\mathbb{E}[R^2\mid x,S]\,
W(x,S)^2
\chi^2\!\left(
\pi(\cdot\mid x,S)\,\|\,\beta(\cdot\mid x,S)
\right)
\right],
\label{eq:ordering-gap}
\end{equation}
where $\pi(\cdot\mid x,S)$ and $\beta(\cdot\mid x,S)$ are the
conditional distributions over the $K!$ orderings of the same
unordered slate. The gap is exactly the ordering-nuisance variance:
it is large when target and behavior agree on the unordered slate only
after averaging over substantially different generation orders.

\paragraph{Algorithm and complexity.}
Algorithm~\ref{alg:forward-dp} computes $F_\mu^x(S_K)$ by visiting all
subsets of the logged slate in increasing size.

\begin{algorithm}[t]
\caption{Forward-DP for one joint slate propensity}
\label{alg:forward-dp}
\begin{algorithmic}[1]
\REQUIRE Slate $S_K\subseteq[M]$, $|S_K|=K$; context $x$; set-sufficient policy $\mu$
\ENSURE $F_\mu^x(S_K)=\mu(S_K\mid x)$
\STATE $F_\mu^x(\emptyset)\leftarrow 1$
\FOR{$t=1,\ldots,K$}
    \FOR{each $S\subseteq S_K$ with $|S|=t$}
        \STATE $F_\mu^x(S)\leftarrow
        \sum_{a\in S}F_\mu^x(S\setminus\{a\})
        \mu(a\mid x,S\setminus\{a\})$
    \ENDFOR
\ENDFOR
\RETURN $F_\mu^x(S_K)$
\end{algorithmic}
\end{algorithm}

The rooted sub-DAG has $2^K$ nodes. If the normalized transition
probabilities $\mu(a\mid x,S)$ are already available, the arithmetic
cost is $O(K\cdot 2^K)$ and memory is $O(2^K)$. For an autoregressive
transformer logger, obtaining $\mu(\cdot\mid x,S)$ requires one
next-item softmax over the remaining catalog per queried subset, giving
\[
O((M+K)\cdot 2^K)
\]
time, dominated by $O(M\cdot 2^K)$ when $M\gtrsim K$.

\begin{proposition}[Subset-query lower bound]
\label{prop:lower-bound}
Fix $M>K$ and a target slate $S_K\subseteq[M]$ with $|S_K|=K$. Any
deterministic algorithm that exactly computes $\mu(S_K\mid x)$ for all
set-sufficient policies using only oracle queries
$\mu(\cdot\mid x,S)$ must query every proper subset of $S_K$ in the
worst case, and hence requires $\Omega(2^K)$ subset-level queries.
\end{proposition}

The proof is in Appendix~\ref{app:proof-lower-bound}. Thus
Forward-DP is query-optimal up to constant factors in the natural
oracle model.

\paragraph{Comparison to alternatives.}
Naively evaluating the unordered propensity requires summing over
$K!$ orderings. Gumbel-top-$K$ sampling
\citep{kool2019stochastic} gives a Monte Carlo approximation whose
accuracy depends on the sampling budget. Ryser's $O(K\cdot 2^K)$
permanent-style subset DP \citep{ryser1963} computes the joint slate
propensity for vanilla Plackett--Luce in the same complexity class as
Forward-DP, but assumes that the per-step softmax does not depend on
the picked prefix. Marginalized-PL closed forms
\citep{ma2021partitioned} likewise apply when
item scores are fixed after observing the context, and target
rank-placement marginals rather than the joint set probability.
Stochastic beam search gives ordered samples without joint
propensities. The gap addressed here is the exact unordered-slate
propensity for context-dependent set-sufficient autoregressive
loggers, where the softmax denominator changes with the partial
slate and Ryser-style fixed-matrix permanents no longer apply.

\paragraph{Forward-DP as a primitive for slate OPE.}
Forward-DP supplies the joint slate propensity consumed by downstream
slate-OPE estimators. We use this primitive in two ways: as the IS
weight $F_\pi(S)/F_\beta(S)$ for the forward-flow estimators (FF-OIS,
FF-WIS, FF-DR) of \S\ref{sec:method-tree}, and as the building block
for context-dependent extensions of two prior slate-OPE constructions.
The marginalized-PL formula \citep{ma2021partitioned} multiplies
inclusion-marginal ratios
$\prod_{d\in S}\pi_{\mathrm{incl}}(d\mid x)/
\beta_{\mathrm{incl}}(d\mid x)$; for vanilla Plackett--Luce these
marginals admit a closed form, but for context-dependent set-sufficient
policies they must be computed from the Forward-DP flow table as
$\sum_S F(S)\mathbf{1}\{d\in S\}$. We refer to this combination as
DP-MPL.
Similarly, the OPCB construction of \citet{kiyohara2024slate}
marginalizes joint propensities over a learnable kth-element
equivalence class; for context-dependent slate loggers, these joint
propensities are supplied by Forward-DP. We refer to this combination
as DP-OPCB. DP-MPL and DP-OPCB are the Forward-DP-enabled extensions of
\citep{ma2021partitioned, kiyohara2024slate} to context-dependent
set-sufficient loggers.

\section{Experiments}
\label{sec:experiments}

We evaluate forward-flow IS on two finite-horizon MDPs and Forward-DP
on KuaiRec slate OPE. The MDP experiments test variance reduction
without fitting a transition model; the slate experiments test exact
unordered-slate propensity computation for context-dependent
autoregressive loggers. We report RMSE for value estimation and
top-$1$, Spearman correlation, and regret for model selection. Full
details, estimator definitions, and additional sweeps are in
Appendix~\ref{app:exp-details}.

\subsection{Finite-horizon MDPs}
\label{sec:exp-mdp}

We evaluate FF-IS on the Sepsis simulator
\citep{oberst2019counterfactual} and ICU-Sepsis
\citep{komorowski2018ai}. In both benchmarks, forward flows are
computed from logged trajectories alone via
Proposition~\ref{prop:quotient-rn}.

\begin{table}[!htbp]
\caption{Finite-horizon MDP validation. Flows use logged trajectories
only. Bold marks the lowest RMSE within each benchmark.
$^{\dagger}$Neural GenDICE's saddle-point optimization did not converge
(defined as $|V - V^\pi| > 1$) in 18/50 Sepsis and 14/50 ICU-Sepsis
trials; reported statistics are on the 32 and 36 converged trials.
Full 50-trial statistics are in Appendix~\ref{app:exp-details}.}
\label{tab:mdp-validation}
\centering
\small
\setlength{\tabcolsep}{4pt}
\renewcommand{\arraystretch}{0.95}
\scalebox{0.90}{%
\begin{tabular}{llrrr}
\toprule
Benchmark & Estimator & Bias & Std & $\downarrow$RMSE \\
\midrule
\multirow{8}{*}{Sepsis}
& OIS        & $+0.039$ & $2.14$    & $2.14$ \\
& WIS        & $+0.105$ & $0.272$   & $0.291$ \\
& FF-OIS     & $+0.001$ & $0.0672$  & $0.0672$ \\
& \textbf{FF-WIS} & $+0.012$ & $0.0555$  & $\mathbf{0.0568}$ \\
& LS         & $+0.012$ & $0.0941$  & $0.0949$ \\
& STAR       & $+0.124$ & $0.0250$  & $0.126$ \\
& DualDICE   & $+0.264$ & $0.004$   & $0.264$ \\
& GenDICE$^{\dagger}$ & $+0.132$ & $0.229$ & $0.264$ \\
\midrule
\multirow{8}{*}{ICU-Sepsis}
& OIS        & $+0.032$ & $0.410$   & $0.412$ \\
& WIS        & $+0.034$ & $0.0409$  & $0.0532$ \\
& FQE        & $-0.322$ & $0.0178$  & $0.323$ \\
& DR         & $-0.286$ & $0.203$   & $0.351$ \\
& FF-OIS     & $-0.005$ & $0.0488$  & $0.0491$ \\
& \textbf{FF-WIS} & $+0.004$ & $0.0262$  & $\mathbf{0.0265}$ \\
& DualDICE   & $-0.759$ & $0.007$   & $0.759$ \\
& GenDICE$^{\dagger}$ & $-0.579$ & $0.430$ & $0.721$ \\
\bottomrule
\end{tabular}%
}
\end{table}

The MDP results are consistent with the variance-reduction intuition
behind quotient conditioning. On Sepsis, FF-WIS reduces RMSE from
\(0.291\) to \(0.0568\); on ICU-Sepsis, it reduces RMSE from
\(0.0532\) to \(0.0265\). These gains use only logged trajectories and
do not rely on a fitted transition model. We include DICE-family
estimators as reference density-ratio baselines; they are not tailored
to these finite-horizon non-stationary benchmarks and are
bias-dominated in Table~\ref{tab:mdp-validation}. Implementation
details and full 50-trial statistics are in
Appendix~\ref{app:exp-details}.

\subsection{KuaiRec slate propensity and OPE}
\label{sec:exp-kuairec}

We next evaluate the slate specialization on KuaiRec
\citep{gao2022kuairec}, using a context-dependent Plackett--Luce
logger with pool size $M=15$ and slate sizes $K\in\{4,6,8\}$.
Table~\ref{tab:kuairec-combined} reports exact propensity runtime and
downstream OPE RMSE.

\begin{table}[!htbp]
\caption{KuaiRec slate experiments: exact propensity runtime and
downstream OPE RMSE. Bold marks the better RMSE within each
trajectory/forward-flow pair.}
\label{tab:kuairec-combined}
\centering
\small
\setlength{\tabcolsep}{3.2pt}
\renewcommand{\arraystretch}{0.95}
\scalebox{0.90}{%
\begin{tabular}{c r r r r r r r r}
\toprule
& \multicolumn{4}{c}{Joint propensity time (s)}
& \multicolumn{4}{c}{Downstream OPE RMSE} \\
\cmidrule(lr){2-5}\cmidrule(lr){6-9}
$K$ & $\binom{M}{K}$ & $K!$ enum & Forward-DP & Gumbel-MC
& OIS & FF-OIS & DR & FF-DR \\
\midrule
4 & $1{,}365$ & $6.2$      & $0.36$ & $26.9$
  & $3.36$  & $\mathbf{1.27}$ & $1.66$ & $\mathbf{0.617}$ \\
6 & $5{,}005$ & $1{,}010$  & $2.93$ & $29.6$
  & $11.46$ & $\mathbf{2.19}$ & $7.81$ & $\mathbf{1.08}$  \\
8 & $6{,}435$ & $97{,}108$ & $8.94$ & $105$
  & $12.30$ & $\mathbf{1.75}$ & $2.02$ & $\mathbf{0.949}$ \\
\bottomrule
\end{tabular}%
}
\end{table}

\begin{figure}[!htbp]
\centering
\includegraphics[width=0.8\linewidth]{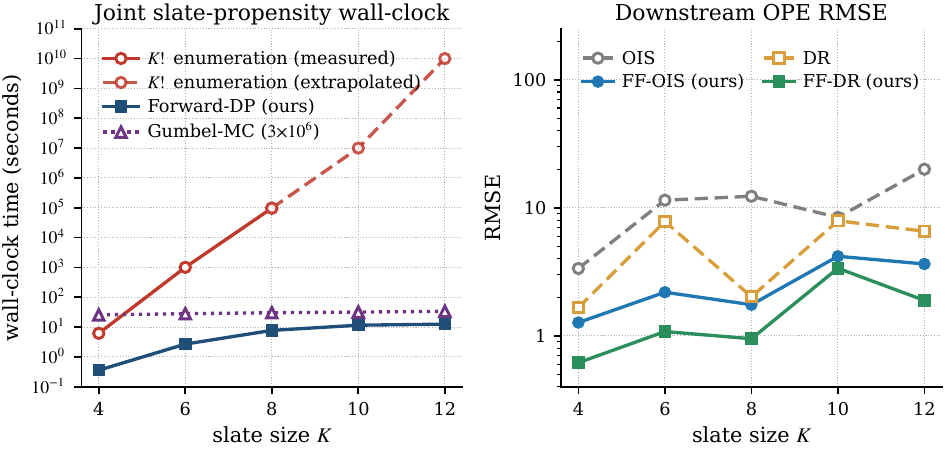}
\caption{
KuaiRec scaling. Left: joint slate-propensity wall-clock time for
\(K!\)-enumeration, Forward-DP, and Gumbel-MC as slate size \(K\)
increases; enumeration is extrapolated beyond \(K=8\), and Gumbel-MC
uses \(3\times10^6\) samples. Forward-DP avoids factorial enumeration
while remaining exact. Right: downstream OPE RMSE for trajectory
weights (OIS, DR) versus forward-flow weights (FF-OIS, FF-DR), using
\(n_{\rm users}=300\) and \(n_{\rm trials}=200\). Within each \(K\),
the forward-flow variants reduce RMSE relative to their
trajectory-weighted counterparts.
}
\vspace{-0.6em}
\label{fig:kuairec-scaling}
\end{figure}

Forward-DP avoids factorial enumeration while remaining exact: at
$K=8$, enumeration takes $97{,}108$ seconds, versus $8.94$ seconds
for Forward-DP. The same trend extends to larger slate sizes: at
$K=10$ and $K=12$ on the same 5-user timing pool, Forward-DP completes
in $12.5$ and $13.7$ seconds respectively, while $K!$-enumeration is
no longer feasible
(extrapolated wall times of $\sim 10^7$ and $\sim 10^{10}$ seconds).
Figure~\ref{fig:kuairec-scaling} plots the wall-clock comparison on a
log-log scale and the corresponding downstream RMSE.
Downstream, FF-OIS and FF-DR consistently improve over OIS and DR
across all $K$. Comparisons to fixed-score Ryser and marginalized-PL alternatives,
together with details on the $K$-sweep, Gumbel-MC setting, and
within-$K$ RMSE interpretation, are given in Appendix~\ref{app:exp-details}.

\subsection{Off-policy model selection}
\label{sec:exp-transformer-ms}

Finally, we apply Forward-DP to off-policy model selection among
autoregressive transformer recommenders. All evaluated policies expose
a set-sufficient next-item interface, so Forward-DP supplies
unordered-slate propensities without architecture-specific
modification. Table~\ref{tab:transformer-ms-K4} reports the main
$K=4$ workflow.

\begin{table}[!htbp]
\caption{Off-policy model selection on KuaiRec at $K=4$ and
$N_{\log}=1000$. Rows are grouped by propensity source; full estimator
definitions are in Appendix~\ref{app:exp-details}.}
\label{tab:transformer-ms-K4}
\centering
\small
\setlength{\tabcolsep}{5pt}
\renewcommand{\arraystretch}{0.95}
\scalebox{0.90}{%
\begin{tabular}{lrrr}
\toprule
Estimator & $\uparrow$Top-1 & $\uparrow$Spearman $\rho$ & $\downarrow$Regret \\
\midrule
\multicolumn{4}{l}{\emph{Trajectory IS}} \\
Tree-OIS    & $0.06$ & $-0.53$ & $0.209$ \\
Tree-WIS    & $0.26$ & $+0.24$ & $0.059$ \\
\midrule
\multicolumn{4}{l}{\emph{Forward-flow IS}} \\
FF-OIS      & $0.08$ & $-0.45$ & $0.204$ \\
FF-WIS      & $0.22$ & $+0.27$ & $0.058$ \\
FF-DR       & $0.18$ & $+0.35$ & $0.072$ \\
\midrule
\multicolumn{4}{l}{\emph{Forward-DP-enabled}} \\
Tree-DR     & $\mathbf{0.28}$ & $\mathbf{+0.42}$ & $0.056$ \\
DP-MPL-OIS  & $0.10$ & $-0.53$ & $0.369$ \\
DP-MPL-WIS  & $0.26$ & $+0.22$ & $0.075$ \\
DP-OPCB-OIS & $0.02$ & $+0.09$ & $0.090$ \\
DP-OPCB-DR  & $0.08$ & $+0.38$ & $\mathbf{0.051}$ \\
\bottomrule
\end{tabular}%
}
\end{table}
The best rows all use Forward-DP-supplied propensities or expectations,
although no estimator dominates all metrics. Tree-DR gives the best
top-$1$ and Spearman scores, while DP-OPCB-DR gives the lowest regret.
The cleanest head-to-head comparison is Tree-OIS versus FF-OIS:
FF-OIS improves on all three metrics. Additional $K$-sweeps and
cross-architecture results are in Appendix~\ref{app:exp-details}.

\section{Related work}
\label{sec:related}

\paragraph{OPE, conditional IS, and marginalization.}
Our quotient-DAG view organizes IS by rollout-tree ratio placement:
trajectories for trajectory IS, prefixes for per-decision IS
\citep{precup2000eligibility,mahmood2014weighted}; conditional
expectations for conditional IS
\citep{rowland2020conditional,liu2020understanding}; the state--time
quotient for finite-horizon MIS
\citep{xie2019towards,liu2020understanding}; and coarser sufficient
quotients $(\phi(s_t),t)$ for abstraction-based OPE
\citep{pavse2023scaling,chaudhari2024neural,majumdar2024conceptdriven}.
DR \citep{jiang2016doubly,thomas2016data} adds control variates, while
DICE \citep{nachum2019dualdice,zhang2020gendice,uehara2020minimax}
estimates discounted-occupancy ratios; DualDICE and GenDICE serve as
\S\ref{sec:exp-mdp} baselines despite a stationary/finite-horizon
target mismatch.

\paragraph{Slate OPE, combinatorial DPs, and slate generators.}
Slate OPE requires the logging propensity of the reward/estimator slate
representation
\citep{swaminathan2017off,vlassis2021control,kiyohara2023off,
kiyohara2024slate,shimizu2024effective}: $\beta(S\mid x)$ for
unordered rewards, but only sampled-order probabilities from
autoregressive loggers. Context-free Plackett--Luce formulas,
Ryser-style subset DPs, and Gumbel-top-$K$ sampling
\citep{ma2021partitioned,ryser1963,kool2019stochastic} solve adjacent
fixed-score or sampling problems, not the context-dependent unordered
joint propensity. Forward-DP supplies this primitive and enables
DP-MPL/DP-OPCB; score-then-mask rankers are set-sufficient, whereas
order-conditioned generative recommenders require picked-set
canonicalization
\citep{kang2018sasrec,sun2019bert4rec,gao2023neuralpgrank,
geng2022p5,cui2022m6rec,rajput2023tiger,sun2023rankgpt,
pradeep2023rankzephyr}.

\section{Conclusion}
\label{sec:conclusion}

We presented Forward-Flow Importance Sampling, a Rao--Blackwellization
of per-decision IS on the layered DAG obtained by quotienting the
rollout tree under any sufficient equivalence relation. The
conditional importance weight admits a closed form as the ratio of
target and behavior forward flows, computable by a single forward
sweep on the quotient DAG. For autoregressive slate generation under
set-sufficiency, the permutation-orbit quotient yields a subset DAG on
which Forward-DP computes the joint slate propensity exactly in
$O((M+K)\cdot 2^K)$ time, fixed-parameter tractable in $K$ and
practical for the $K\le 10$ slate sizes typical of slate-OPE
deployments. For context-dependent autoregressive Plackett--Luce
loggers whose next-item softmax depends on the partial slate, this
closes the joint-propensity gap and supplies a missing OPE primitive
for transformer-based slate recommenders. The framework recovers MIS
at the $(\text{state},\text{time})$ quotient and slate IS at the
permutation-orbit quotient; experiments on Sepsis, ICU-Sepsis,
KuaiRec, and off-policy model selection for transformer recommenders
validate both the framework and the slate primitive.

Future directions include continuous-state extensions via learned
partitions or kernels, bias analysis for $\epsilon$-sufficient
quotients, and forward-flow weights combined with doubly-robust
baselines.

\bibliographystyle{plainnat}
\bibliography{refs}

\appendix

\section{Proofs and additional details for Section~\ref{sec:method}}
\label{app:method-proofs}

\subsection{Proof of Proposition~\ref{prop:quotient-rn}}
\label{app:proof-quotient-rn}

Fix a layer $t$ and a quotient class $z\in Z_t$ with
$F_\beta(z)>0$. By the history-level support assumption, for any
pre-action history $h_t$ with $P_\beta(h_t)>0$ the prefix likelihood
ratio is
\[
\rho_{1:t-1}(h_t)
=
\frac{P_\pi(h_t)}{P_\beta(h_t)},
\]
because the target and behavior policies share the same initial-state,
transition, and reward kernels, and differ only in their action
probabilities; for histories with $P_\beta(h_t)=0$ the assumption
forces $P_\pi(h_t)=0$ as well, so such histories contribute zero mass
to both measures. By Bayes' rule, for $h_t\in z$ with $P_\beta(h_t)>0$,
\[
P_\beta(h_t\mid z_t=z)
=
\frac{P_\beta(h_t)}{F_\beta(z)}.
\]
Therefore, restricting the conditional expectation to the
$P_\beta$-support inside $z$,
\[
\mathbb{E}_\beta[\rho_{1:t-1}\mid z_t=z]
=
\sum_{\substack{h_t\in z\\P_\beta(h_t)>0}}
\frac{P_\beta(h_t)}{F_\beta(z)}
\frac{P_\pi(h_t)}{P_\beta(h_t)}
=
\frac{1}{F_\beta(z)}
\sum_{h_t\in z}P_\pi(h_t)
=
\frac{F_\pi(z)}{F_\beta(z)},
\]
where the second equality uses $P_\pi(h_t)=0$ on
$\{h_t\in z:P_\beta(h_t)=0\}$ to extend the sum back to all of $z$.
Since $P_{\mu,t}^{Z}(z)=F_\mu(z)$ by definition of the pushforward
measure, the same identity gives
\[
\frac{dP_{\pi,t}^{Z}}{dP_{\beta,t}^{Z}}(z)
=
\frac{P_{\pi,t}^{Z}(z)}{P_{\beta,t}^{Z}(z)}
=
\frac{F_\pi(z)}{F_\beta(z)}
\]
on any finite quotient class with $F_\beta(z)>0$. The general
countable case follows identically by replacing sums over atoms with
the corresponding Radon--Nikodym derivative of the pushed-forward
measures.

\subsection{Forward-flow IS as a per-term conditional estimator}
\label{app:ff-per-term}

We justify the per-decision statement used in
\S\ref{sec:method-tree}. The $t$th PDIS summand is
\[
X_t
=
\gamma^{t-1}\rho_{1:t-1}(h_t)\rho_t(h_t,a_t)r_t.
\]
Under a sufficient quotient, the current action distribution is
class-measurable:
\[
\rho_t(h_t,a_t)=\rho_t(z_t,a_t)
=
\frac{\pi(a_t\mid z_t)}{\beta(a_t\mid z_t)}.
\]
Moreover, sufficiency implies that, under the behavior policy, the
conditional distribution of $(a_t,r_t)$ given $h_t$ depends on $h_t$
only through $z_t$. Hence conditioning on $(z_t,a_t,r_t)$ does not
change the conditional distribution over histories inside the quotient
class beyond conditioning on $z_t$ itself. Therefore, by
Proposition~\ref{prop:quotient-rn},
\[
\mathbb{E}_\beta[
\rho_{1:t-1}(h_t)\mid z_t,a_t,r_t]
=
\mathbb{E}_\beta[
\rho_{1:t-1}(h_t)\mid z_t]
=
\frac{F_\pi(z_t)}{F_\beta(z_t)}.
\]
It follows that
\[
\mathbb{E}_\beta[X_t\mid z_t,a_t,r_t]
=
\gamma^{t-1}
\frac{F_\pi(z_t)}{F_\beta(z_t)}
\rho_t(z_t,a_t)r_t,
\]
which is exactly the $t$th FF-IS summand. Thus each FF-IS summand is a
conditional expectation of the corresponding PDIS summand. Consequently
it is unbiased for the same per-decision target, and by the law of
total variance,
\[
\mathrm{Var}_\beta\!\left(
\mathbb{E}_\beta[X_t\mid z_t,a_t,r_t]
\right)
\le
\mathrm{Var}_\beta(X_t).
\]
This establishes the per-term Rao--Blackwell property. It does not
imply a variance ordering for the sum
$\sum_t X_t$, because the covariance terms between different time steps
may change after conditioning.

\subsection{Proof of Proposition~\ref{prop:variance-gap}}
\label{app:proof-variance-gap}

Let $G(\tau)=g(q(\tau))$ and
\[
w(q)=
\frac{dP_\pi^q}{dP_\beta^q}(q).
\]
By the law of total variance,
\[
\mathrm{Var}_\beta(\rho G)
=
\mathrm{Var}_\beta\!\left(
\mathbb{E}_\beta[\rho G\mid q]
\right)
+
\mathbb{E}_\beta\!\left[
\mathrm{Var}_\beta(\rho G\mid q)
\right].
\]
Since $G=g(q)$ is $q$-measurable and
$\mathbb{E}_\beta[\rho\mid q]=w(q)$ is the standard pushforward
Radon--Nikodym identity (i.e., for $\rho=dP_\pi/dP_\beta$ we have
$\mathbb{E}_\beta[\rho\mid q]=dP_\pi^q/dP_\beta^q(q)=w(q)$, the same
push-forward computation as in
Proposition~\ref{prop:quotient-rn} but at the terminal quotient
rather than at a prefix layer),
\[
\mathbb{E}_\beta[\rho G\mid q]
=
g(q)\mathbb{E}_\beta[\rho\mid q]
=
w(q)g(q).
\]
Therefore
\[
\mathrm{Var}_\beta(\rho G)
-
\mathrm{Var}_\beta(w(q)g(q))
=
\mathbb{E}_\beta\!\left[
\mathrm{Var}_\beta(\rho G\mid q)
\right].
\]
Again using $G=g(q)$,
\[
\mathrm{Var}_\beta(\rho G\mid q)
=
g(q)^2\mathrm{Var}_\beta(\rho\mid q),
\]
which proves
\[
\mathrm{Var}_\beta(\rho G)
-
\mathrm{Var}_\beta(w(q)g(q))
=
\mathbb{E}_\beta\!\left[
g(q)^2\mathrm{Var}_\beta(\rho\mid q)
\right].
\]

For a finite quotient space, fix a quotient class $z$. For
$q(\tau)=z$,
\[
\rho(\tau)
=
\frac{P_\pi(\tau)}{P_\beta(\tau)}
=
\frac{P_\pi(z)P_\pi(\tau\mid z)}
     {P_\beta(z)P_\beta(\tau\mid z)}
=
w(z)\frac{P_\pi(\tau\mid z)}{P_\beta(\tau\mid z)}.
\]
Thus
\[
\mathrm{Var}_\beta(\rho\mid z)
=
w(z)^2
\mathrm{Var}_{\tau\sim P_\beta(\cdot\mid z)}
\!\left[
\frac{P_\pi(\tau\mid z)}{P_\beta(\tau\mid z)}
\right].
\]
Since the likelihood ratio
$P_\pi(\tau\mid z)/P_\beta(\tau\mid z)$ has expectation $1$ under
$P_\beta(\cdot\mid z)$, the variance is exactly the chi-square
divergence:
\[
\mathrm{Var}_{P_\beta(\cdot\mid z)}
\!\left[
\frac{P_\pi(\tau\mid z)}{P_\beta(\tau\mid z)}
\right]
=
\chi^2\!\left(
P_\pi(\cdot\mid z)\,\|\,P_\beta(\cdot\mid z)
\right).
\]
Multiplying by $P_\beta(q=z)=F_\beta(z)$ and summing over $z$ gives
\[
\sum_z
F_\beta(z)g(z)^2w(z)^2
\chi^2\!\left(
P_\pi(\cdot\mid z)\,\|\,P_\beta(\cdot\mid z)
\right).
\]

\subsection{Proof of Proposition~\ref{prop:flow-exact}}
\label{app:proof-flow-exact}

We prove the claim by induction on $|S|$. For the empty set,
\[
F_\mu^x(\emptyset)=1=\mu(\emptyset\mid x).
\]
Assume the claim holds for all subsets of size less than $|S|$, and
consider a nonempty set $S$. Every ordering of $S$ has a unique last
selected item $a\in S$. Conversely, for every $a\in S$, any ordering
of $S\setminus\{a\}$ followed by $a$ gives an ordering of $S$ whose
last item is $a$. Therefore the unordered propensity can be partitioned
by the last item:
\[
\mu(S\mid x)
=
\sum_{a\in S}
\mu(S\setminus\{a\}\mid x)
\mu(a\mid x,S\setminus\{a\}),
\]
where set-sufficiency is used to make the probability of selecting
$a$ depend only on the picked set $S\setminus\{a\}$ rather than on the
order used to reach it. By the induction hypothesis,
\[
\mu(S\setminus\{a\}\mid x)=F_\mu^x(S\setminus\{a\}),
\]
so
\[
\mu(S\mid x)
=
\sum_{a\in S}
F_\mu^x(S\setminus\{a\})
\mu(a\mid x,S\setminus\{a\})
=
F_\mu^x(S).
\]
This proves the proposition.

\subsection{Derivation of the ordering-nuisance variance gap}
\label{app:ordering-gap}

We derive Eq.~\eqref{eq:ordering-gap}. Fix a context $x$ and unordered
slate $S$. Let $\sigma$ denote the latent generation ordering of $S$.
The trajectory-style ordered estimator uses the weight
\[
\rho_{\mathrm{ord}}(x,\sigma)
=
\frac{\pi(\sigma\mid x)}{\beta(\sigma\mid x)},
\]
whereas the DAG estimator uses the unordered weight
\[
W(x,S)
=
\frac{\pi(S\mid x)}{\beta(S\mid x)}.
\]
Conditioning on $(x,S)$, define the conditional ordering
distributions
\[
\pi(\sigma\mid x,S)
=
\frac{\pi(\sigma\mid x)}{\pi(S\mid x)},
\qquad
\beta(\sigma\mid x,S)
=
\frac{\beta(\sigma\mid x)}{\beta(S\mid x)}.
\]
For any ordering $\sigma$ of $S$,
\[
\rho_{\mathrm{ord}}(x,\sigma)
=
W(x,S)
\frac{\pi(\sigma\mid x,S)}{\beta(\sigma\mid x,S)}.
\]
Therefore
\[
\mathbb{E}_{\beta(\cdot\mid x,S)}
[\rho_{\mathrm{ord}}(x,\sigma)]
=
W(x,S),
\]
and
\[
\mathrm{Var}_{\beta(\cdot\mid x,S)}
(\rho_{\mathrm{ord}}(x,\sigma))
=
W(x,S)^2
\chi^2\!\left(
\pi(\cdot\mid x,S)\,\|\,\beta(\cdot\mid x,S)
\right).
\]

Now assume the reward distribution depends on $(x,S)$ but not on the
ordering, and is conditionally independent of $\sigma$ given $(x,S)$.
Conditioning on $(x,S,R)$, the DAG estimator is deterministic:
$W(x,S)R$, while the ordered estimator is
$\rho_{\mathrm{ord}}(x,\sigma)R$. Their conditional means agree:
\[
\mathbb{E}_\beta[
\rho_{\mathrm{ord}}(x,\sigma)R\mid x,S,R]
=
W(x,S)R.
\]
Hence the variance removed by replacing the ordered estimator with the
DAG estimator is the expected conditional variance
\[
\mathbb{E}\!\left[
R^2
\mathrm{Var}_{\beta(\cdot\mid x,S)}
(\rho_{\mathrm{ord}}(x,\sigma))
\right].
\]
Substituting the chi-square expression above and averaging over
$x\sim p_X$ and $S\sim\beta(\cdot\mid x)$ gives
\[
\mathbb{E}_{x\sim p_X,S\sim\beta(\cdot\mid x)}
\!\left[
\mathbb{E}[R^2\mid x,S]\,
W(x,S)^2
\chi^2\!\left(
\pi(\cdot\mid x,S)\,\|\,\beta(\cdot\mid x,S)
\right)
\right],
\]
which is Eq.~\eqref{eq:ordering-gap}.

\subsection{Proof of Proposition~\ref{prop:lower-bound}}
\label{app:proof-lower-bound}

We prove the lower bound in the subset-oracle model used in
Proposition~\ref{prop:lower-bound}. The algorithm may query an oracle
for transition distributions of the form $\mu(\cdot\mid x,S)$ at
subsets $S\subseteq[M]$. It must output the exact unordered propensity
$\mu(S_K\mid x)$ for every set-sufficient policy $\mu$.

Fix $M>K$ and a target slate $S_K$ with $|S_K|=K$. Suppose a
deterministic algorithm does not query some proper subset
$U\subset S_K$. We construct two set-sufficient policies that return
identical answers to every oracle query made by the algorithm but have
different values of $\mu(S_K\mid x)$.

Because $M>K$, there exists an item $b\in[M]\setminus S_K$. Since
$U\subset S_K$ is proper, there exists an item
$a\in S_K\setminus U$. Consider two policies that agree on all queried
subsets and on all subsets except $U$. At the unqueried subset $U$,
let the first policy assign probability mass $p+\delta$ to item $a$
and probability mass $q-\delta$ to item $b$, while the second assigns
probability $p$ to $a$ and $q$ to $b$, with $\delta>0$ chosen small
enough that all probabilities remain nonnegative. All other transition
probabilities at $U$ are adjusted identically so that both
distributions remain normalized.

The two policies are indistinguishable to the algorithm because they
differ only at the unqueried subset $U$. However, they assign different
total mass to paths that pass through $U$ and then continue inside
$S_K$. Shifting probability from $b\notin S_K$ to
$a\in S_K\setminus U$ changes the probability of eventually completing
the exact slate $S_K$, because choosing $b$ makes it impossible to end
with unordered slate $S_K$ of size $K$. Therefore the two policies have
different values of $\mu(S_K\mid x)$ while producing identical oracle
responses on all queried subsets.

Thus any exact deterministic algorithm must query every proper subset
of $S_K$ in the worst case. There are $2^K-1$ such subsets, giving an
$\Omega(2^K)$ subset-query lower bound. Forward-DP queries each proper
subset once, and is therefore query-optimal up to constant factors in
this oracle model.

\section{Additional discussion of related estimators}
\label{app:related-estimators}

The quotient likelihood-ratio identity used in the main text is a
special case of conditional importance sampling. In the notation of
Section~\ref{sec:method}, choosing the conditioning variable to be the
full pre-action history \(h_t\) gives ordinary per-decision IS, since
\[
\frac{F_\pi(h_t)}{F_\beta(h_t)}
=
\frac{P_\pi(h_t)}{P_\beta(h_t)}
=
\rho_{1:t-1}.
\]
Choosing the quotient variable \(z_t=(s_t,t)\) in a Markov state space
gives finite-horizon marginalized IS. More generally, choosing
\(z_t=(\phi(s_t),t)\) for a sufficient abstraction \(\phi\) gives
known-abstraction MIS. The slate construction uses a different
history-prefix quotient: all ordered prefixes that contain the same
picked set are merged. The common object in all cases is the
pushforward likelihood ratio on the quotient space.

The algorithmic distinction in the slate setting is that the quotient
density itself is not readily available from an autoregressive logger.
Naively computing the unordered propensity requires summing over
\(K!\) orderings. Gumbel-top-\(K\) sampling gives a Monte Carlo
approximation, while marginalized Plackett--Luce closed forms apply to
fixed-score settings where item scores do not change with the partial
slate. Forward-DP instead handles context-dependent set-sufficient
loggers, where the next-item softmax may change after every picked set.

\paragraph{Finite-horizon IS variants.}
We first spell out the finite-horizon estimators to clarify the
relationship between ordinary IS, weighted IS, per-decision IS, and
forward-flow IS. For a trajectory
\(\tau=(s_1,a_1,r_1,\ldots,s_H,a_H,r_H)\), define
\[
G(\tau)=\sum_{t=1}^H\gamma^{t-1}r_t,
\qquad
\rho_t(\tau)=\frac{\pi(a_t\mid s_t)}{\beta(a_t\mid s_t)},
\qquad
\rho_{1:t}(\tau)=\prod_{u=1}^t \rho_u(\tau).
\]
The ordinary trajectory IS estimator is
\[
\widehat V_{\mathrm{OIS}}(\pi)
=
\frac{1}{N}\sum_{i=1}^N
\rho_{1:H}(\tau^{(i)})G(\tau^{(i)}).
\]
The self-normalized or weighted variant is
\[
\widehat V_{\mathrm{WIS}}(\pi)
=
\frac{
\sum_{i=1}^N \rho_{1:H}(\tau^{(i)})G(\tau^{(i)})
}{
\sum_{i=1}^N \rho_{1:H}(\tau^{(i)})
}.
\]
WIS generally has lower weight instability than OIS but is biased in
finite samples.

The per-decision IS estimator uses the prefix ratio only up to the
reward time:
\[
\widehat V_{\mathrm{PDIS}}(\pi)
=
\frac{1}{N}\sum_{i=1}^N
\sum_{t=1}^H
\gamma^{t-1}
\rho_{1:t}(\tau^{(i)})r_t^{(i)} .
\]
Under a sufficient quotient, the prefix ratio
\(\rho_{1:t-1}\) can be replaced by its conditional expectation within
the quotient class. This gives the forward-flow per-decision estimator
\[
\widehat V_{\mathrm{FF}}(\pi)
=
\frac{1}{N}\sum_{i=1}^N
\sum_{t=1}^H
\gamma^{t-1}
\frac{F_\pi(z_t^{(i)})}{F_\beta(z_t^{(i)})}
\frac{\pi(a_t^{(i)}\mid z_t^{(i)})}
     {\beta(a_t^{(i)}\mid z_t^{(i)})}
r_t^{(i)} .
\]
When the quotient flows are known, this estimator is unbiased under the
same support condition as PDIS, and each summand is a Rao--Blackwellized
version of the corresponding PDIS summand. As discussed in the main
text, this is a per-term statement and does not imply a variance
ordering for the full multi-step sum.

A self-normalized per-decision variant can also be formed by normalizing
the weights at each decision time:
\[
\widehat V_{\mathrm{WPDIS}}(\pi)
=
\sum_{t=1}^H
\gamma^{t-1}
\frac{
\sum_{i=1}^N \rho_{1:t}(\tau^{(i)})r_t^{(i)}
}{
\sum_{i=1}^N \rho_{1:t}(\tau^{(i)})
},
\]
with the analogous forward-flow version replacing
\(\rho_{1:t-1}\) by \(F_\pi(z_t)/F_\beta(z_t)\). In the experiments, we
use the notation FF-OIS and FF-WIS for the ordinary and self-normalized
forms after this quotient-weight replacement.

\paragraph{Slate-level ordered and quotient weights.}
For slate OPE, the logged action may be an ordered generation sequence
\(\sigma=(\sigma_1,\ldots,\sigma_K)\), while the reward depends only on
the unordered slate \(S=\{\sigma_1,\ldots,\sigma_K\}\). The ordered
trajectory-tree ratio is
\[
\rho_{\mathrm{tree}}(x,\sigma)
=
\frac{\pi(\sigma\mid x)}{\beta(\sigma\mid x)}
=
\prod_{t=1}^K
\frac{
\pi(\sigma_t\mid x,h_{<t})
}{
\beta(\sigma_t\mid x,h_{<t})
}.
\]
This is the weight used by Tree-OIS and Tree-WIS. In contrast, the
quotient or forward-flow slate ratio is
\[
W_{\mathrm{FF}}(x,S)
=
\frac{\pi(S\mid x)}{\beta(S\mid x)}
=
\frac{F_\pi^x(S)}{F_\beta^x(S)} ,
\]
where \(F_\mu^x(S)\) is computed exactly by Forward-DP under
set-sufficiency. This is the weight used by FF-OIS, FF-WIS, and FF-DR.

The ordered ordinary IS estimator is
\[
\widehat V_{\mathrm{Tree\text{-}OIS}}(\pi)
=
\frac{1}{N}\sum_{n=1}^N
\rho_{\mathrm{tree}}(x_n,\sigma_n)R_n,
\]
whereas the quotient ordinary IS estimator is
\[
\widehat V_{\mathrm{FF\text{-}OIS}}(\pi)
=
\frac{1}{N}\sum_{n=1}^N
W_{\mathrm{FF}}(x_n,S_n)R_n .
\]
Their self-normalized variants are
\[
\widehat V_{\mathrm{Tree\text{-}WIS}}(\pi)
=
\frac{
\sum_{n=1}^N
\rho_{\mathrm{tree}}(x_n,\sigma_n)R_n
}{
\sum_{n=1}^N
\rho_{\mathrm{tree}}(x_n,\sigma_n)
},
\]
and
\[
\widehat V_{\mathrm{FF\text{-}WIS}}(\pi)
=
\frac{
\sum_{n=1}^N
W_{\mathrm{FF}}(x_n,S_n)R_n
}{
\sum_{n=1}^N
W_{\mathrm{FF}}(x_n,S_n)
}.
\]
Thus ``Tree'' denotes weighting on the ordered autoregressive rollout
tree, while ``FF'' denotes weighting after pushing the policy
distributions forward to the unordered slate quotient.

\paragraph{Doubly robust slate estimators.}
Let \(\hat r(x,S)\) be a fixed or independently fitted reward model.
The direct-method value for a context \(x\) is
\[
\widehat m_\pi(x)
=
\sum_{S:|S|=K}
\pi(S\mid x)\hat r(x,S)
=
\sum_{S:|S|=K}
F_\pi^x(S)\hat r(x,S).
\]
When the candidate slate space is small enough, this sum is evaluated
exactly over all size-\(K\) subsets; otherwise it can be approximated
using the same Forward-DP primitive on the queried slate family.

The forward-flow doubly robust estimator is
\[
\widehat V_{\mathrm{FF\text{-}DR}}(\pi)
=
\frac{1}{N}\sum_{n=1}^N
\left[
\widehat m_\pi(x_n)
+
W_{\mathrm{FF}}(x_n,S_n)
\{R_n-\hat r(x_n,S_n)\}
\right].
\]
For comparison, the ordered Tree-DR variant uses the same direct-method
term but corrects the residual with the ordered trajectory ratio:
\[
\widehat V_{\mathrm{Tree\text{-}DR}}(\pi)
=
\frac{1}{N}\sum_{n=1}^N
\left[
\widehat m_\pi(x_n)
+
\rho_{\mathrm{tree}}(x_n,\sigma_n)
\{R_n-\hat r(x_n,S_n)\}
\right].
\]
Tree-DR and FF-DR therefore differ only in the residual correction:
Tree-DR corrects by the probability of the realized generation order,
whereas FF-DR corrects by the probability of the unordered slate. With
exact propensities and a fixed or cross-fitted reward model, the DR
estimator is unbiased if either the propensity model is correct or the
reward model is correct. Self-normalized and plug-in variants may
introduce finite-sample bias, as usual.

\paragraph{Forward-DP-enabled extensions.}
Forward-DP also supplies the probability tables needed by existing
slate-OPE constructions whose original forms assume access to slate
propensities or fixed-score Plackett--Luce marginals. In DP-MPL, we
compute the inclusion marginal of item \(d\) under policy \(\mu\) as
\[
p_{\mu,\mathrm{incl}}(d\mid x)
=
\sum_{S:|S|=K}
F_\mu^x(S)\mathbf{1}\{d\in S\},
\]
and use these marginals inside the marginalized-PL weighting rule. In
DP-OPCB, Forward-DP supplies the joint slate propensities that are then
marginalized over the OPCB equivalence classes. We use the prefixes
``DP-MPL'' and ``DP-OPCB'' rather than ``FF-MPL'' or ``FF-OPCB'' because
these estimators use Forward-DP as a propensity-computation primitive,
not the forward-flow slate ratio \(F_\pi(S)/F_\beta(S)\) directly.

\section{Empirical quotient-flow estimation in the MDP experiments}
\label{app:empirical-flow-estimation}

The theoretical statements in Section~\ref{sec:method-tree} assume that the
quotient forward flows, or equivalently the quotient likelihood ratios
$F_\pi(z)/F_\beta(z)$, are known. In the finite-horizon MDP experiments of
Section~\ref{sec:exp-mdp}, we do not fit a transition model and do not use
oracle transition probabilities. Instead, we estimate quotient ratios directly
from the logged trajectories using the conditional-expectation identity in
Proposition~\ref{prop:quotient-rn}.

For a quotient class $z\in Z_t$, let
\[
\mathcal I_t(z)=\{i: z_t(\tau^{(i)})=z\}
\]
be the indices of logged trajectories whose layer-$t$ pre-action history lies
in class $z$. The empirical quotient ratio used in the plug-in estimator is
\begin{equation}
\widehat w_t(z)
=
\frac{1}{|\mathcal I_t(z)|}
\sum_{i\in\mathcal I_t(z)}
\rho_{1:t-1}(\tau^{(i)}),
\qquad |\mathcal I_t(z)|>0.
\label{eq:empirical-flow-ratio}
\end{equation}
This is the sample analogue of
$\mathbb E_\beta[\rho_{1:t-1}\mid z_t=z]=F_\pi(z)/F_\beta(z)$.
The resulting plug-in FF estimator replaces $F_\pi(z_t^{(i)})/F_\beta(z_t^{(i)})$
in Eq.~\eqref{eq:ff-is} by $\widehat w_t(z_t^{(i)})$.

This empirical construction is useful for validating quotient conditioning from
logged data, but it should not be confused with the oracle-flow estimator in the
theory. If the same trajectories are used both to estimate
$\widehat w_t(z)$ and to evaluate the IS summands, the resulting estimator can
have finite-sample plug-in bias. A sample-split version avoids this coupling by
estimating $\widehat w_t$ on one split and evaluating Eq.~\eqref{eq:ff-is} on an
independent split. A leave-one-out variant similarly defines
\begin{equation}
\widehat w_t^{(-i)}(z)
=
\frac{1}{|\mathcal I_t(z)|-1}
\sum_{j\in\mathcal I_t(z),\,j\ne i}
\rho_{1:t-1}(\tau^{(j)}),
\label{eq:loo-flow-ratio}
\end{equation}
whenever $|\mathcal I_t(z)|>1$. The MDP experiments are intended as validation
of the quotient-weighting mechanism; the main algorithmic contribution in the
slate setting computes the relevant quotient flows exactly by Forward-DP.

Empty or singleton quotient classes are handled in the usual support-limited
way. If a class is never observed under the behavior data, it cannot contribute
to the empirical estimator. If a leave-one-out denominator is zero, the
corresponding sample is omitted from the leave-one-out diagnostic or replaced by
the split-estimated ratio. These implementation choices affect only the
empirical MDP validation, not the exact slate Forward-DP results.

\section{Additional implementation details for Forward-DP}
\label{app:forward-dp-implementation}

\paragraph{Bit-mask implementation.}
For a logged slate $S_K=\{d_1,\ldots,d_K\}$, Forward-DP is implemented over
bit masks $m\in\{0,1\}^K$. A mask $m$ represents the subset
$S(m)=\{d_j:m_j=1\}$. Masks are visited in increasing Hamming weight. For each
nonempty mask $m$, the update is
\begin{equation}
F_\mu^x(m)
=
\sum_{j:m_j=1}
F_\mu^x(m\setminus\{j\})
\mu(d_j\mid x,S(m\setminus\{j\})).
\label{eq:mask-forward-dp}
\end{equation}
The output for the logged slate is $F_\mu^x(\mathbf 1)$, where $\mathbf 1$ is
the all-one mask.

\paragraph{Caching next-item probabilities.}
For each proper subset $U\subset S_K$, the implementation queries the policy
once for the next-item distribution $\mu(\cdot\mid x,U)$ and caches the
probabilities of the still-unselected logged items $S_K\setminus U$. If the
policy is a neural generator whose logits must be recomputed after each partial
slate, this query includes the forward pass and the softmax over the remaining
catalog. If normalized probabilities are already available, only the cached
transition probabilities are needed.

\paragraph{Numerical stability.}
For the slate sizes considered in the experiments, ordinary floating-point
summation is sufficient. For larger $K$ or sharper policies, the same recursion
can be run in log space:
\begin{equation}
\log F_\mu^x(S)
=
\operatorname{logsumexp}_{a\in S}
\left\{
\log F_\mu^x(S\setminus\{a\})+
\log \mu(a\mid x,S\setminus\{a\})
\right\}.
\label{eq:log-forward-dp}
\end{equation}
This does not change the query complexity. Zero-probability transitions are
represented as $-\infty$ in log space. The slate-level support condition used
by Eq.~\eqref{eq:dag-slate-is} requires $F_\beta^x(S)>0$ whenever
$F_\pi^x(S)>0$ for the logged evaluation distribution.

\paragraph{Computing full distributions and marginals.}
Algorithm~\ref{alg:forward-dp} computes the propensity of one logged slate by
restricting the subset DAG to subsets of that slate. Some downstream estimators
or diagnostics require quantities over all size-$K$ slates in a candidate pool.
In that case the same recursion can be run over the full subset lattice
$\{S\subseteq[M]: |S|\le K\}$. The cost is
\begin{equation}
O\!\left(M\sum_{\ell=0}^{K-1}{M\choose \ell}
+
K{M\choose K}\right)
\end{equation}
when each queried subset requires a softmax over the catalog, and the resulting
flows give both joint slate probabilities $F_\mu^x(S)$ and inclusion marginals
\begin{equation}
\mu_{\mathrm{incl}}(d\mid x)
=
\sum_{S\subseteq[M]: |S|=K,\,d\in S}
F_\mu^x(S).
\label{eq:inclusion-marginal-dp}
\end{equation}
The logged-slate version used for propensity correction is cheaper because it
requires only the $2^K$ subsets of the observed slate.

\section{Estimator definitions used in model selection}
\label{app:model-selection-estimators}

This section spells out the estimators used in Table~\ref{tab:transformer-ms-K4}.
For a logged example $(x_n,\sigma_n,S_n,R_n)$, let
\[\rho_{\mathrm{ord},n}=\pi(\sigma_n\mid x_n)/\beta(\sigma_n\mid x_n)\]
be the ordered trajectory ratio and
\[W_n=F_\pi^{x_n}(S_n)/F_\beta^{x_n}(S_n)\]
be the unordered forward-flow slate ratio.

\paragraph{Tree-OIS and Tree-WIS.}
Tree-OIS uses the ordered ratio,
\begin{equation}
\widehat V_{\mathrm{Tree\text{-}OIS}}(\pi)
=
\frac{1}{N}\sum_{n=1}^N \rho_{\mathrm{ord},n}R_n.
\end{equation}
Tree-WIS self-normalizes the same ordered weights,
\begin{equation}
\widehat V_{\mathrm{Tree\text{-}WIS}}(\pi)
=
\frac{\sum_{n=1}^N \rho_{\mathrm{ord},n}R_n}
{\sum_{n=1}^N \rho_{\mathrm{ord},n}}.
\end{equation}
These estimators do not require Forward-DP.

\paragraph{FF-OIS and FF-WIS.}
FF-OIS replaces the ordered trajectory ratio by the unordered quotient ratio,
\begin{equation}
\widehat V_{\mathrm{FF\text{-}OIS}}(\pi)
=
\frac{1}{N}\sum_{n=1}^N W_n R_n,
\end{equation}
and FF-WIS self-normalizes the same weights,
\begin{equation}
\widehat V_{\mathrm{FF\text{-}WIS}}(\pi)
=
\frac{\sum_{n=1}^N W_nR_n}{\sum_{n=1}^N W_n}.
\end{equation}
These are the direct head-to-head estimators for ordered versus unordered
weighting.

\paragraph{FF-DR and Tree-DR.}
Let $\widehat q(x,S)$ denote a fixed or independently fitted slate reward model.
The FF-DR estimator is
\begin{equation}
\widehat V_{\mathrm{FF\text{-}DR}}(\pi)
=
\frac{1}{N}\sum_{n=1}^N
\left[
\sum_{S:|S|=K}F_\pi^{x_n}(S)\widehat q(x_n,S)
+
W_n\{R_n-\widehat q(x_n,S_n)\}
\right].
\label{eq:ff-dr-app}
\end{equation}
Tree-DR uses the same direct-method expectation but retains the ordered residual
weight:
\begin{equation}
\widehat V_{\mathrm{Tree\text{-}DR}}(\pi)
=
\frac{1}{N}\sum_{n=1}^N
\left[
\sum_{S:|S|=K}F_\pi^{x_n}(S)\widehat q(x_n,S)
+
\rho_{\mathrm{ord},n}\{R_n-\widehat q(x_n,S_n)\}
\right].
\label{eq:tree-dr-app}
\end{equation}
Thus Tree-DR still uses Forward-DP for the direct-method expectation, even
though its residual correction is trajectory-weighted.

\paragraph{DP-MPL.}
DP-MPL uses Forward-DP to compute inclusion marginals in the context-dependent
setting. For each policy $\mu$ and context $x$, define
$\mu_{\mathrm{incl}}(d\mid x)$ as in Eq.~\eqref{eq:inclusion-marginal-dp}. The
MPL-style weight for a logged slate is
\begin{equation}
W_{\mathrm{MPL}}(x,S)
=
\prod_{d\in S}
\frac{\pi_{\mathrm{incl}}(d\mid x)}
{\beta_{\mathrm{incl}}(d\mid x)}.
\label{eq:dp-mpl-weight}
\end{equation}
DP-MPL-OIS and DP-MPL-WIS use this weight in ordinary and self-normalized IS,
respectively. The estimator is biased in general because the product of
inclusion-marginal ratios is not the joint slate ratio; it is included as a
Forward-DP-enabled extension of marginalized-PL-style weighting.

\paragraph{DP-OPCB.}
DP-OPCB follows the OPCB abstraction idea by grouping slates into learned or
specified equivalence classes and using Forward-DP-supplied joint propensities
to compute class-level probabilities. If $c(S)$ denotes the OPCB class of slate
$S$, define
\begin{equation}
P_\mu(c\mid x)
=
\sum_{S: |S|=K,\, c(S)=c}F_\mu^x(S).
\end{equation}
The OPCB class weight is $P_\pi(c(S_n)\mid x_n)/P_\beta(c(S_n)\mid x_n)$.
DP-OPCB-OIS uses this class weight directly, while DP-OPCB-DR adds the
corresponding direct-method/control-variate term. Because OPCB intentionally
coarsens the action space, its objective can trade bias for variance; this is
why its lowest-regret row in Table~\ref{tab:transformer-ms-K4} need not have the
highest exact top-$1$ accuracy.

\section{Limitations and scope}
\label{app:limitations}

\paragraph{Sufficiency of the quotient.}
The finite-horizon FF-IS construction assumes that the quotient is sufficient in
the sense of Definition~\ref{def:sufficient}. Coarser reward-compatible
partitions can still be useful, but exact unbiasedness requires that the
continuation law relevant to the estimator be compatible with the quotient.
For example, a reward-class quotient alone is not sufficient unless histories in
the same class also have compatible future transitions, actions, and rewards
under the policies being compared.

\paragraph{Set-sufficiency in slate generation.}
Forward-DP is exact only for the stochastic next-item interface in
Definition~\ref{def:set-sufficiency}. A transformer or LLM architecture does
not automatically satisfy this property. The deployed policy must present the
picked items through a canonical set representation, or otherwise ensure that
previously picked items affect the next-item distribution only through the picked
set. If the model conditions on the emitted order, then the unordered slate
propensity remains a sum over $K!$ distinct order-conditioned paths unless the
interface is changed.
Canonicalization is a deployment-time policy choice, not an OPE-method
limitation: we evaluate whichever set-sufficient policy is deployed, and
non-canonicalized order-conditioned variants are simply different policies, for
which exact propensity computation requires either $K!$ enumeration or
approximate Monte Carlo regardless of the OPE estimator used.

\paragraph{Empirical set-sufficiency violation.}
To quantify how far from set-sufficient an off-the-shelf transformer
recommender actually is, we measure, for random picked subsets
$S\subseteq[M]$ with $|S|=t$, the next-item distribution under each
ordering of $S$ and report the median (across user/subset draws) of
the maximum pairwise total-variation distance over orderings.
Set-sufficiency holds exactly iff this TVD is identically zero; any
non-zero value bounds the bias incurred by Forward-DP if the
canonicalization step is omitted.
Table~\ref{tab:set-suff-tvd} reports this diagnostic for the
canonicalized SASRec and Qwen2.5-3B reranker deployments used in
\S\ref{sec:exp-transformer-ms}.

\begin{table}[H]
\caption{Empirical violation of set-sufficiency under non-canonicalized
prefix orderings on KuaiRec ($M=15$, item-letter prompts).
Two subset-sampling distributions are compared: \emph{random} draws
$S\subseteq[M]$ uniformly; \emph{$\beta$-induced} draws $S$ as the
prefix at step $|S|$ of a canonicalized $\beta$ trajectory of length
$K_{\max}\!=\!8$. Each row reports the median, 90th percentile, and
mean (over user/subset draws) of the maximum pairwise TVD across
orderings of $S$. Set-sufficiency requires all entries to be exactly
$0$; the larger entries on $\beta$-induced rows therefore quantify
the bias the OPE estimator actually encounters under the deployed
behavior policy. Each cell uses 160 (user, subset) draws.}
\label{tab:set-suff-tvd}
\centering
\small
\setlength{\tabcolsep}{5pt}
\begin{tabular}{l r r r r r r r}
\toprule
& $|S|$
& \multicolumn{3}{c}{SASRec (KuaiRec, $M=15$)}
& \multicolumn{3}{c}{Qwen2.5-3B (KuaiRec, $M=15$)} \\
\cmidrule(lr){3-5}\cmidrule(lr){6-8}
& & med.\,max & $p_{90}$\,max & med.\,mean
& med.\,max & $p_{90}$\,max & med.\,mean \\
\midrule
\multirow{7}{*}{\rotatebox{90}{random}}
& $1$ & $0.00$  & $0.00$  & $0.00$  & $0.00$  & $0.00$  & $0.00$  \\
& $2$ & $0.15$  & $0.28$  & $0.15$  & $0.04$  & $0.07$  & $0.04$  \\
& $3$ & $0.14$  & $0.23$  & $0.10$  & $0.07$  & $0.11$  & $0.04$  \\
& $4$ & $0.22$  & $0.34$  & $0.12$  & $0.11$  & $0.16$  & $0.05$  \\
& $5$ & $0.21$  & $0.34$  & $0.11$  & $0.14$  & $0.19$  & $0.07$  \\
& $6$ & $0.19$  & $0.29$  & $0.09$  & $0.15$  & $0.24$  & $0.07$  \\
& $7$ & $0.28$  & $0.43$  & $0.14$  & $0.16$  & $0.26$  & $0.08$  \\
\midrule
\multirow{7}{*}{\rotatebox{90}{$\beta$-induced}}
& $1$ & $0.00$  & $0.00$  & $0.00$  & $0.00$  & $0.00$  & $0.00$  \\
& $2$ & $0.20$  & $0.32$  & $0.20$  & $0.04$  & $0.09$  & $0.04$  \\
& $3$ & $0.19$  & $0.30$  & $0.12$  & $0.08$  & $0.11$  & $0.05$  \\
& $4$ & $0.25$  & $0.40$  & $0.14$  & $0.12$  & $0.17$  & $0.06$  \\
& $5$ & $0.25$  & $0.39$  & $0.13$  & $0.15$  & $0.22$  & $0.08$  \\
& $6$ & $0.27$  & $0.31$  & $0.12$  & $0.18$  & $0.26$  & $0.09$  \\
& $7$ & $0.38$  & $0.45$  & $0.17$  & $0.21$  & $0.28$  & $0.11$  \\
\bottomrule
\end{tabular}
\end{table}

The two recommenders show distinct but moderate violations on the
random-subset rows: on Qwen2.5-3B the median maximum pairwise TVD
grows from $0$ at $|S|=1$ to $\approx 0.16$ at $|S|=7$, while on
SASRec the same quantity grows to $\approx 0.28$ ($p_{90}\!\approx\!0.43$),
reflecting the stronger positional dependence of an autoregressive
sequence model relative to a prompt-based reranker. The
$\beta$-induced rows are uniformly larger than the matching random
rows: at $|S|=7$, SASRec rises from median TVD $0.28$ to $0.38$, and
Qwen from $0.16$ to $0.21$. This direction of effect is expected.
The behavior policy preferentially samples high-score items together,
which are exactly the configurations for which the next-item softmax
is most concentrated (and therefore most order-sensitive); a
diagnostic restricted to uniformly random subsets understates the
violation that the OPE estimator actually encounters at deployment.
A worst-case bound on the resulting joint-propensity error is obtained
by partitioning $\mu_a(S\mid x)=\sum_{a\in S}\mu_a(S\setminus\{a\}\mid x)
\bar\mu_a(a\mid x,S\setminus\{a\})$ where
$\bar\mu_a(a\mid x,S\setminus\{a\})$ is the average of
$\mu_a(a\mid x,\sigma')$ over orderings $\sigma'$ of $S\setminus\{a\}$
under $\mu_a$; per-step TVD $\le\varepsilon$ gives
$|\bar\mu_a(\cdot\mid x,T)-\mu_c(\cdot\mid x,T)|_{\mathrm{TV}}\le\varepsilon$
for every prefix subset $T$. Writing
$D_\ell=\max_{|S|=\ell}|F_c^x(S)-\mu_a(S\mid x)|$, the triangle
inequality applied to this averaged recursion yields
\(D_\ell\le \ell\,D_{\ell-1}+2\varepsilon\,\ell\), and hence
$D_K\le 2\varepsilon\,K!\,\sum_{j=0}^{K-1}1/j!=O(K!\,\varepsilon)$.
This bound is exponential in $K$, and at the empirical TVDs in
Table~\ref{tab:set-suff-tvd} ($\varepsilon\approx 0.1$--$0.3$, $K\le 7$)
it is large enough to be vacuous in absolute terms; equivalently, the
per-ordering deviation can be at most $\sim K\varepsilon$ but summing
over the $K!$ orderings of $S$ gives no useful absolute bound at this
scale. The substantive point of the table is therefore not a tight
bias number but a structural one: the deployed policy and the
canonical set-sufficient policy are slightly different objects, the
OPE estimator evaluates whichever is presented to it through the
interface, and the canonicalization in
\S\ref{sec:exp-transformer-ms} is part of the target-policy
specification, not a transparent preprocessing step. A tighter
data-dependent bias bound for non-canonicalized deployments is left to
future work.

\paragraph{Reward invariance.}
The slate variance-gap statement assumes that the reward distribution depends on
$(x,S)$ and not on the latent generation ordering. If the displayed order itself
affects user response, then the relevant evaluation object is an ordered slate or
a deterministic display transformation of the unordered set. In that case the
quotient should be defined with respect to the representation on which the
reward is actually measured.

\paragraph{Scaling with slate size.}
Forward-DP removes the factorial dependence on $K$ but remains exponential in
$K$. This is appropriate for moderate slate sizes, where $2^K$ is feasible and
$K!$ is not, but it is not intended for very large slates without additional
approximation or structure.

\paragraph{Known stochastic interface and support.}
Like other propensity-based OPE estimators, the slate estimators require access
to the behavior propensity on the evaluated representation and a support
condition. If the logging system does not expose calibrated next-item
probabilities, or if the target places mass on slates with zero behavior
propensity, exact Forward-DP computation alone cannot remove the resulting OPE
identification problem.

\section{Experiment details}
\label{app:exp-details}

\paragraph{Reproducibility.}
All experiments use single-CPU runs with random seed $42$. The Sepsis
simulator uses $N=5000$ logged trajectories and 500 trials; ICU-Sepsis
uses 200 trials. KuaiRec downstream OPE uses $N=500$ and 200 trials.
Model-selection experiments use 50 trials with logged budgets
$N_{\mathrm{log}}\in\{50,100,200,500,1000\}$.

\paragraph{Sepsis simulator.}
The Sepsis simulator of \citet{oberst2019counterfactual} is a
stochastic MDP with $720$ ground states, $8$ actions, horizon $H=5$,
and terminal rewards $\{+1,0,-1\}$ for discharge, non-terminal, and
death respectively. The target policy is
$\pi=\mathrm{softmax}(Q^*)$ at temperature $0.1$, and the behavior
policy is $\beta=\epsilon$-greedy$(Q^*)$ at $\epsilon=0.3$.
The ground-truth value is $V(\pi)=-0.2683$. The learned-abstraction
baselines in Table~\ref{tab:mdp-validation} are LS
\citep{pavse2023scaling} and STAR \citep{chaudhari2024neural}.

\paragraph{DICE baselines.}
DualDICE \citep{nachum2019dualdice} and GenDICE
\citep{zhang2020gendice} are run with $\gamma=0.99$ and the canonical
estimator $V = \mathbb{E}_{(s,a)\sim d_\beta}[\zeta(s,a)\,r(s,a)] /
(1-\gamma)$ on each estimator's primal density-ratio output.
Both are parameterized as a 2-layer MLP with 64 hidden units (state
and action embeddings of dimension 64) and trained for $20{,}000$
steps (DualDICE) or $15{,}000$ steps (GenDICE) with Adam, batch size
2048, and gradient clipping at norm $5$. DualDICE uses the primal
form ($\zeta(s,a)=\partial f(\mathrm{bellman})$ for $f(x)=x^2/2$);
GenDICE uses the saddle-point form with
$\tau\geq 0$ via softplus, learning rates
$(\eta_\tau,\eta_f)=(10^{-4},10^{-3})$, normalization penalty
$\lambda=10$, and chi-square penalty $\alpha=1$. Each estimator is
run for 50 trials with seeds $\{43,\ldots,92\}$. The reported
Table~\ref{tab:mdp-validation} entries are bias, std, and RMSE
relative to the undiscounted ground truth $V^\pi$.

The full 50-trial GenDICE bias / std / RMSE before any trimming are
$(-11.78,\,58.16,\,59.34)$ on Sepsis and
$(+96.32,\,359.40,\,372.08)$ on ICU-Sepsis: a long tail of seeds for
which the saddle-point optimization places density-ratio mass on
states with non-zero terminal reward, blowing up
$\mathbb{E}_\beta[\tau\,r]$ and through the $1/(1-\gamma)$ scale,
$V$. Discounted-occupancy DICE estimators target the stationary
$\gamma$-discounted distribution under an infinite-horizon MDP; on
the finite-horizon non-stationary Sepsis benchmarks, neither the
infinite-horizon decomposition $d_\pi^\gamma=(1-\gamma)\sum_t\gamma^t
\Pr_\pi(s_t=\cdot,a_t=\cdot)$ nor the saddle's stationary-distribution
ansatz match the data-generating process, and the saddle is highly
sensitive to seed. We define non-convergence as
$|V-V^\pi|>1$ (loose enough that any reasonable IPS-family RMSE
remains converged) and report the trimmed bias / std / RMSE on
converged trials in Table~\ref{tab:mdp-validation}, marked with
$^{\dagger}$. The median estimator value across all 50 trials, robust
to the tail, is $-0.069$ on Sepsis (median bias $+0.199$, MAD
$0.155$) and $+0.206$ on ICU-Sepsis (median bias $-0.628$, MAD
$0.200$). DualDICE on the same protocol does not exhibit this
divergence (50/50 trials within a $|V-V^\pi|=1$ ball) but its bias
remains an order of magnitude larger than FF-WIS.

For the Table~\ref{tab:mdp-validation} entries, $10{,}000$-bootstrap
$95\%$ CIs over the $50$-trial sample are: DualDICE Sepsis bias
$[+0.263,+0.265]$, RMSE $[0.264,0.266]$; DualDICE ICU-Sepsis bias
$[-0.762,-0.758]$, RMSE $[0.758,0.762]$; GenDICE$^{\dagger}$ Sepsis
(trimmed $32/50$) bias $[+0.046,+0.206]$, RMSE $[0.216,0.316]$;
GenDICE$^{\dagger}$ ICU-Sepsis (trimmed $36/50$) bias
$[-0.703,-0.425]$, RMSE $[0.677,0.762]$. DualDICE CIs are tight
because the saddle is well-conditioned in our setting; GenDICE CIs
are wider despite trimming, reflecting residual variability among
``converged'' trials.

\begin{figure}[H]
\centering
\includegraphics[width=\linewidth]{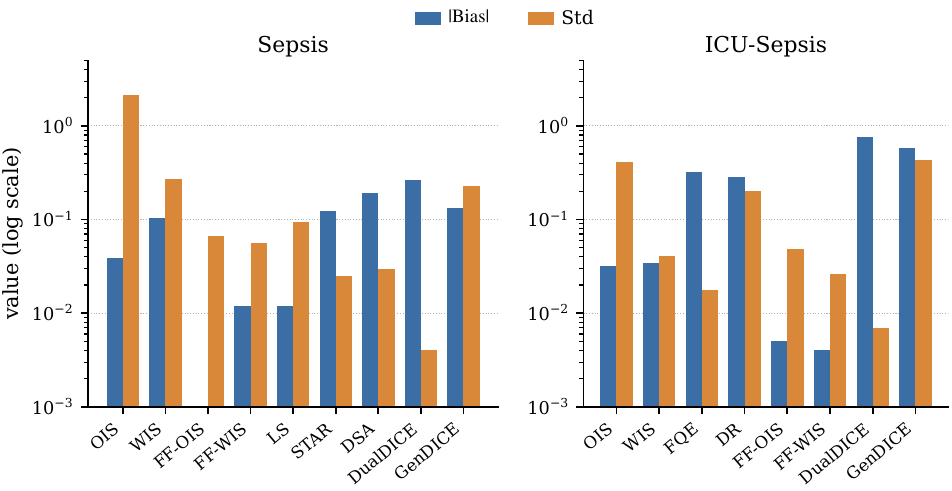}
\caption{Bias versus standard-deviation decomposition of RMSE on the
finite-horizon MDP benchmarks (visualizing
Table~\ref{tab:mdp-validation}). Bars on log scale. STAR on Sepsis
and FQE on ICU-Sepsis are bias-dominated; OIS is
variance-dominated on both benchmarks; the DICE-family estimators
(DualDICE, GenDICE) are bias-dominated on both, with bias an order of
magnitude larger than FF-WIS; FF-OIS and FF-WIS are small on both
axes. GenDICE bars use trimmed-trial statistics
(see Table~\ref{tab:mdp-validation}).}
\label{fig:mdp-bias-std}
\end{figure}

\paragraph{ICU-Sepsis.}
ICU-Sepsis~\citep{komorowski2018ai} is a tabular MDP with $716$
states and $25$ actions fit from the MIMIC-III intensive-care
database. Target-policy rollouts are used only to approximate
$V(\pi)=0.833$ as ground truth and never enter the estimator.
Figure~\ref{fig:mdp-bias-std} decomposes the
Table~\ref{tab:mdp-validation} RMSE entries into bias and standard
deviation, showing that learned-abstraction baselines (STAR on
Sepsis; FQE on ICU-Sepsis) are bias-dominated, while OIS is
variance-dominated; FF-OIS and FF-WIS are small on both axes.

\paragraph{KuaiRec slate setup.}
KuaiRec~\citep{gao2022kuairec} provides a $1{,}411\times 3{,}327$
user--video matrix at $99.6\%$ density. We construct a per-item ranking
task with pool size $M=15$, slate sizes $K\in\{4,6,8\}$, and a
context-dependent Plackett--Luce behavior policy implemented as
per-item softmax sampling without replacement. The reward is the
geometric mean of true watch ratios plus Gaussian noise
$\sigma=1$, with $n_{\mathrm{users}}=300$.

\paragraph{KuaiRec scaling details for Figure~\ref{fig:kuairec-scaling}.}
Figure~\ref{fig:kuairec-scaling} evaluates how slate size \(K\) affects
both joint-propensity computation and downstream OPE. The left panel
compares exact enumeration over all \(K!\) generation orders, Forward-DP,
and a Gumbel-top-\(K\) Monte Carlo estimator
\citep{kool2019stochastic}. Direct \(K!\)-enumeration is measured up to
\(K=8\) and extrapolated for \(K>8\), where direct measurement is
computationally infeasible. Gumbel-MC uses \(3\times 10^6\) samples per
user. In the KuaiRec timing experiment the behavior policy is a
fixed-score Plackett--Luce logger, so Gumbel-top-\(K\) sampling admits a
vectorized fast path. This makes Gumbel-MC competitive in wall-clock
time on this particular plain-PL timing benchmark, but it remains an
approximation and does not provide exact propensities.

The distinction from Forward-DP becomes important for the
context-dependent loggers used in the model-selection experiments.
Fixed-score Plackett--Luce and Gumbel-top-\(K\) samplers assume that the
item scores are fixed after observing the context
\citep{ma2021partitioned,kool2019stochastic}.
They therefore do not directly compute the unordered joint propensity
when the next-item softmax is recomputed after each partial slate, as in
the SASRec and Qwen logger interfaces of Section~\ref{sec:exp-transformer-ms}.
Forward-DP instead queries the policy on each picked subset and is exact
for the stated set-sufficient context-dependent interface.

\paragraph{\(K\)-sweep RMSE summary.}
The right panel of Figure~\ref{fig:kuairec-scaling} reports downstream
OPE RMSE for \(K\in\{4,6,8,10,12\}\) with
\(n_{\mathrm{users}}=300\) and \(n_{\mathrm{trials}}=200\). Absolute
RMSE values are not directly comparable across \(K\), because the
ground-truth target value changes substantially with slate size:
\[
V(\pi)=
4.64,\ 3.27,\ 0.05,\ -4.71,\ -10.86
\quad
\text{for}
\quad
K=4,6,8,10,12,
\]
respectively. In addition, the number of size-\(K\) slates,
\(\binom{M}{K}\), peaks around \(K=7\)--\(8\) for \(M=15\), and the
negative pairwise interaction term in the reward construction becomes
dominant at larger \(K\). We therefore interpret the right panel
within each fixed \(K\), comparing each trajectory-weighted estimator to
its forward-flow counterpart.

Within each \(K\), the relative RMSE reductions are stable. Table~\ref{tab:kuairec-rmse-ratio}
reports ratios against the corresponding trajectory-weighted baselines.
FF-OIS reduces RMSE relative to OIS for all tested \(K\), and FF-DR
reduces RMSE relative to trajectory-IS DR for all tested \(K\). The
largest reductions occur when the ordered prefix ratios are most
unstable, especially at \(K=6\) and \(K=12\).

\begin{table}[H]
\caption{Within-\(K\) RMSE ratios on KuaiRec
(\(M=15\), \(n_{\mathrm{users}}=300\),
\(n_{\mathrm{trials}}=200\)). Lower than \(1.00\) indicates lower RMSE
than the denominator. Ratios should be compared within the same \(K\),
not across different \(K\).}
\label{tab:kuairec-rmse-ratio}
\centering
\small
\setlength{\tabcolsep}{8pt}
\begin{tabular}{r r r r r}
\toprule
\(K\) & FF-OIS / OIS & DR / OIS & FF-DR / OIS & FF-DR / DR \\
\midrule
4  & \(0.378\) & \(0.494\) & \(0.184\) & \(0.372\) \\
6  & \(0.191\) & \(0.682\) & \(0.094\) & \(0.138\) \\
8  & \(0.142\) & \(0.164\) & \(0.077\) & \(0.470\) \\
10 & \(0.497\) & \(0.941\) & \(0.401\) & \(0.426\) \\
12 & \(0.182\) & \(0.328\) & \(0.095\) & \(0.288\) \\
\bottomrule
\end{tabular}
\end{table}

\paragraph{Fixed-score propensity baselines.}
We compare Forward-DP to three adjacent fixed-score or sampling
baselines. First, direct enumeration sums the probabilities of all
\(K!\) orderings and is exact but factorial in \(K\). Second,
Gumbel-MC uses the Gumbel-top-\(K\) trick to sample without replacement
from fixed item scores and estimates unordered propensities by Monte
Carlo \citep{kool2019stochastic}. Third, permanent-style subset dynamic
programs such as Ryser's formula \citep{ryser1963} and marginalized
Plackett--Luce closed forms
\citep{ma2021partitioned} exploit fixed-score
structure. These alternatives are relevant comparisons because they
also avoid explicit \(K!\) enumeration in fixed-score settings.

However, they do not target the same quantity as Forward-DP when the
logger is context-dependent at each partial slate. In a re-prompting or
set-conditioned logger, the softmax denominator and item logits may
change after each picked set \(S_{<t}\). Applying a fixed-score Ryser
computation to the re-prompting SASRec logger of
Section~\ref{sec:exp-transformer-ms} gives average distributional
\(L^1\) errors of \(0.39\), \(0.50\), and \(0.57\) at
\(K=4,6,8\), after normalizing the prefix-independent estimates over
all \(\binom{M}{K}\) slates for each user. Forward-DP is designed for
this context-dependent set-sufficient setting: it recomputes or queries
the next-item distribution at every picked subset and returns the exact
unordered propensity for the deployed interface.

\paragraph{Off-policy model selection: setup.}
KuaiRec is used with item pool $M=20$ for $K\in\{4,6\}$ and
$M=15$ for $K=8$ to match the propensity-timing setup of
Section~\ref{sec:exp-kuairec}. User-pool size is
$n_{\mathrm{users}}=300$ for SASRec-only model selection, matching the
OPCB scale of \citet{kiyohara2024slate}, and
$n_{\mathrm{users}}=30$ for cross-architecture runs that include the
LLM, since the LLM logit cache scales linearly with users. SASRec-only
rankings agree at the two user-pool scales. The deployment logger is
$\beta=\text{SASRec}[\text{step }200,T=3.0]$, a weakly trained
high-temperature checkpoint standing in for an exploration-aware
production deployment. Reward is the position-allocated discounted
watch ratio with a pairwise diminishing-returns interaction term
($\lambda=0.1$), following \citet{kiyohara2024slate}.

\paragraph{Off-policy model selection: candidate pool.}
Thirteen target candidates are evaluated jointly in the SASRec-only
setting: six SASRec checkpoints at training step
$\in\{200,400,800,1600,3200,6400\}$ with $T=0.7$; four step-$800$
copies at $T\in\{0.3,0.5,1.0,1.5\}$; two reward-aware references
ScorerPL$[T=0.3]$ and ScorerPL$[T=1.0]$ on the held-out reward
scorer; and a uniform-random generator. Cross-architecture runs add
three LLM-Qwen2.5-3B candidates at $T\in\{0.5,0.7,1.0\}$ prompted with
KuaiRec second-level category captions. Ground-truth $V(\pi_c)$ is
computed exactly by enumerating all $\binom{M}{K}$ subsets and
weighting them with the Forward-DP propensity $F_{\pi_c}$.

\begin{figure}[H]
\centering
\includegraphics[width=\linewidth]{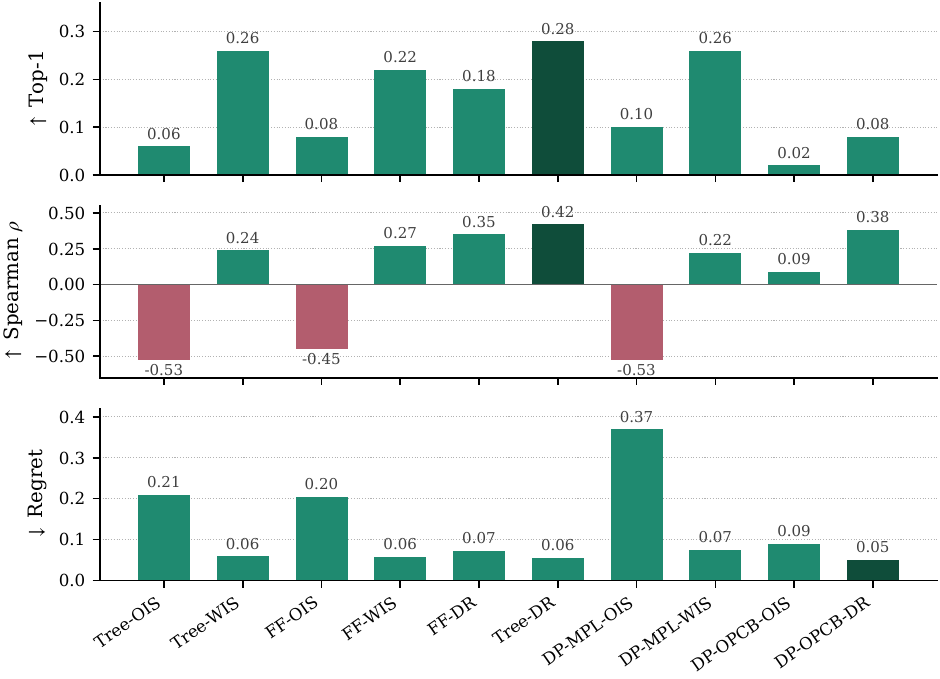}
\caption{$K=4$ off-policy model-selection metrics by estimator
(visualizing Table~\ref{tab:transformer-ms-K4}). Higher is better for
top-$1$ and Spearman; lower is better for regret. Best per panel:
Tree-DR (top-$1$, Spearman) and DP-OPCB-DR (regret), both
Forward-DP-enabled.}
\label{fig:k4-modelselect}
\end{figure}

\paragraph{Estimator definitions for model selection.}
Tree-OIS and Tree-WIS use the trajectory prefix-ratio weight and
require no unordered-slate joint propensity. FF-OIS, FF-WIS, and FF-DR
replace the trajectory prefix ratio by the forward-flow slate weight
$F_\pi(S)/F_\beta(S)$. Tree-DR uses the trajectory-IS residual but
computes the direct-method expectation
\[
V_{\mathrm{DM}}(\pi)=\sum_S F_\pi(S)\widehat q(S)
\]
with Forward-DP. DP-MPL-OIS and DP-MPL-WIS apply the marginalized-PL
construction using inclusion marginals
$\sum_S F(S)\mathbf{1}\{d\in S\}$ supplied by Forward-DP in the
context-dependent setting. DP-OPCB-OIS and DP-OPCB-DR apply the OPCB
construction of \citet{kiyohara2024slate} with Forward-DP-supplied joint
propensities and the SLOPE+ kth-element rule.
Figure~\ref{fig:k4-modelselect} visualizes the per-estimator metrics
of Table~\ref{tab:transformer-ms-K4}.

\paragraph{Additional slate-size sweeps.}
Changing $K$ changes both the ordering-orbit size and the candidate
value landscape. At $K=6$, the SASRec-only candidate values collapse
to a narrow range $[3.28,3.64]$ with $V(\beta)=3.62$ near the top,
making broadness-biased estimators unusually favorable; DP-MPL-OIS
therefore attains top-$1$ accuracy $1.00$, Spearman $+0.91$, and zero
regret in this specific landscape. This dominance is not uniform: at
$K=4$, the same estimator has regret $0.369$. The direct forward-flow
comparison shows the expected growth with the ordering orbit: at
$K=8$, FF-OIS reaches top-$1$ accuracy $0.44$ versus $0.24$ for
Tree-OIS, with $40{,}320$ latent orderings per slate.

\paragraph{Cross-architecture migration.}
Adding three LLM-Qwen2.5-3B candidates yields a $K$-dependent
deployment verdict. At $K=4$, the LLM cluster sits below trained
SASRec checkpoints ($V\in[4.02,4.13]$ versus $[4.49,4.62]$). At
$K=6$, the same LLM cluster moves to the top
($V\in[3.64,3.69]$, beating the best SASRec at $3.61$). At $K=8$ on
$M=15$, SASRec[step $400$, $T=0.7$] wins back with $V=1.04$, while the
LLM cluster is mid-pack at $V\in[0.76,0.84]$. In this offline
benchmark, once the logger interface and reward table are available,
Forward-DP supplies the unordered-slate propensities needed to evaluate
these counterfactual slate-size and architecture choices without new
online interaction.
Table~\ref{tab:cross-arch} summarizes the $K$-dependent verdicts.

\begin{table}[H]
\caption{Cross-architecture migration verdict by slate size $K$.
Values are computed exactly by enumerating all $\binom{M}{K}$
size-$K$ subsets and weighting them with the Forward-DP propensity
$F_{\pi_c}$.}
\label{tab:cross-arch}
\centering
\small
\setlength{\tabcolsep}{6pt}
\begin{tabular}{ccccc}
\toprule
$K$ & $M$ & LLM-Qwen $V$ range & Best SASRec $V$ & Verdict \\
\midrule
4 & 20 & $[4.02, 4.13]$ & $4.62$ (range $[4.49, 4.62]$) & SASRec \\
6 & 20 & $[3.64, 3.69]$ & $3.61$                        & LLM \\
8 & 15 & $[0.76, 0.84]$ & $1.04$ (step $400$, $T=0.7$)  & SASRec \\
\bottomrule
\end{tabular}
\end{table}

\paragraph{DP-OPCB-WIS at $K=4$ SASRec-only.}
The self-normalized DP-OPCB-WIS variant is omitted from
Table~\ref{tab:transformer-ms-K4} because it attains top-$1$ accuracy
$0.00$ across all 50 trials on this benchmark, with Spearman $+0.14$
and regret $0.052$. Inspecting the per-trial argmax shows that
self-normalization combined with the SLOPE+ kth shrinkage of
\citet{kiyohara2024slate} concentrates the predicted-best on a fixed
broader-distribution candidate, SASRec[step $800$, $T=1.0$] with
$V=4.37$, rather than the sharper true-best, SASRec[step $800$,
$T=0.3$] with $V=4.42$. Thus regret remains competitive while exact
top-$1$ accuracy is zero in this landscape.

\section*{NeurIPS Paper Checklist}

\begin{enumerate}

\item {\bf Claims}
    \item[] Question: Do the main claims made in the abstract and introduction accurately reflect the paper's contributions and scope?
    \item[] Answer: \answerYes{}
    \item[] Justification: The two contributions claimed in the abstract and introduction (the rollout-tree to quotient-DAG view of OPE in \S\ref{sec:method-tree} and the Forward-DP exact-propensity algorithm for context-dependent set-sufficient autoregressive slate loggers in \S\ref{sec:method-slate}) are both formalized as propositions with proofs in Appendix~\ref{app:method-proofs} and validated empirically in \S\ref{sec:experiments}.
    \item[] Guidelines:
    \begin{itemize}
        \item The answer \answerNA{} means that the abstract and introduction do not include the claims made in the paper.
        \item The abstract and/or introduction should clearly state the claims made, including the contributions made in the paper and important assumptions and limitations. A \answerNo{} or \answerNA{} answer to this question will not be perceived well by the reviewers.
        \item The claims made should match theoretical and experimental results, and reflect how much the results can be expected to generalize to other settings.
        \item It is fine to include aspirational goals as motivation as long as it is clear that these goals are not attained by the paper.
    \end{itemize}

\item {\bf Limitations}
    \item[] Question: Does the paper discuss the limitations of the work performed by the authors?
    \item[] Answer: \answerYes{}
    \item[] Justification: \S\ref{sec:setup} (``Scope of set-sufficiency'') states for which recommender deployment styles set-sufficiency holds and for which it does not; \S\ref{sec:method-tree} notes that per-decision FF-IS gives a per-term Rao--Blackwell statement that does not imply a full-estimator variance ordering; \S\ref{sec:method-slate} bounds Forward-DP's regime of practicality at $K\le 10$; Appendix~\ref{app:limitations} consolidates the assumptions and scope of the framework (sufficiency, set-sufficiency, reward invariance, scaling with $K$, support and known stochastic interface); the Conclusion lists open directions on continuous-state extensions and approximate sufficiency.
    \item[] Guidelines:
    \begin{itemize}
        \item The answer \answerNA{} means that the paper has no limitation while the answer \answerNo{} means that the paper has limitations, but those are not discussed in the paper.
        \item The authors are encouraged to create a separate ``Limitations'' section in their paper.
        \item The paper should point out any strong assumptions and how robust the results are to violations of these assumptions (e.g., independence assumptions, noiseless settings, model well-specification, asymptotic approximations only holding locally). The authors should reflect on how these assumptions might be violated in practice and what the implications would be.
        \item The authors should reflect on the scope of the claims made, e.g., if the approach was only tested on a few datasets or with a few runs. In general, empirical results often depend on implicit assumptions, which should be articulated.
        \item The authors should reflect on the factors that influence the performance of the approach.
        \item The authors should discuss the computational efficiency of the proposed algorithms and how they scale with dataset size.
        \item If applicable, the authors should discuss possible limitations of their approach to address problems of privacy and fairness.
        \item While the authors might fear that complete honesty about limitations might be used by reviewers as grounds for rejection, a worse outcome might be that reviewers discover limitations that aren't acknowledged in the paper. The authors should use their best judgment and recognize that individual actions in favor of transparency play an important role in developing norms that preserve the integrity of the community. Reviewers will be specifically instructed to not penalize honesty concerning limitations.
    \end{itemize}

\item {\bf Theory assumptions and proofs}
    \item[] Question: For each theoretical result, does the paper provide the full set of assumptions and a complete (and correct) proof?
    \item[] Answer: \answerYes{}
    \item[] Justification: All four propositions (\ref{prop:quotient-rn}, \ref{prop:variance-gap}, \ref{prop:flow-exact}, \ref{prop:lower-bound}) state their assumptions in the proposition body. Full proofs appear in Appendix~\ref{app:method-proofs} (\S\ref{app:proof-quotient-rn}, \S\ref{app:proof-variance-gap}, \S\ref{app:proof-flow-exact}, \S\ref{app:proof-lower-bound}). The variance gap derivation specialized to slate OPE is in \S\ref{app:ordering-gap}.
    \item[] Guidelines:
    \begin{itemize}
        \item The answer \answerNA{} means that the paper does not include theoretical results.
        \item All the theorems, formulas, and proofs in the paper should be numbered and cross-referenced.
        \item All assumptions should be clearly stated or referenced in the statement of any theorems.
        \item The proofs can either appear in the main paper or the supplemental material, but if they appear in the supplemental material, the authors are encouraged to provide a short proof sketch to provide intuition.
        \item Inversely, any informal proof provided in the core of the paper should be complemented by formal proofs provided in appendix or supplemental material.
        \item Theorems and Lemmas that the proof relies upon should be properly referenced.
    \end{itemize}

\item {\bf Experimental result reproducibility}
    \item[] Question: Does the paper fully disclose all the information needed to reproduce the main experimental results of the paper to the extent that it affects the main claims and/or conclusions of the paper (regardless of whether the code and data are provided or not)?
    \item[] Answer: \answerYes{}
    \item[] Justification: All experiments use public benchmarks (Sepsis simulator of \citet{oberst2019counterfactual}, MIMIC-III ICU-Sepsis tabular MDP of \citet{komorowski2018ai}, KuaiRec of \citet{gao2022kuairec}) and the open Qwen2.5-3B model. Appendix~\ref{app:exp-details} specifies behavior and target policies, candidate pool, reward construction with diminishing-returns coefficient, logged-budget grid, single-CPU runs, and random seed $42$.
    \item[] Guidelines:
    \begin{itemize}
        \item The answer \answerNA{} means that the paper does not include experiments.
        \item If the paper includes experiments, a \answerNo{} answer to this question will not be perceived well by the reviewers: Making the paper reproducible is important, regardless of whether the code and data are provided or not.
        \item If the contribution is a dataset and\slash or model, the authors should describe the steps taken to make their results reproducible or verifiable.
        \item Depending on the contribution, reproducibility can be accomplished in various ways. For example, if the contribution is a novel architecture, describing the architecture fully might suffice, or if the contribution is a specific model and empirical evaluation, it may be necessary to either make it possible for others to replicate the model with the same dataset, or provide access to the model.
    \end{itemize}

\item {\bf Open access to data and code}
    \item[] Question: Does the paper provide open access to the data and code, with sufficient instructions to faithfully reproduce the main experimental results, as described in supplemental material?
    \item[] Answer: \answerYes{}
    \item[] Justification: All datasets are publicly available (KuaiRec via official release; Sepsis simulator open source; MIMIC-III via standard PhysioNet credentialing). An anonymized code release accompanying the supplemental material reproduces every figure, table, and propensity-timing experiment.
    \item[] Guidelines:
    \begin{itemize}
        \item The answer \answerNA{} means that paper does not include experiments requiring code.
        \item Please see the NeurIPS code and data submission guidelines (\url{https://neurips.cc/public/guides/CodeSubmissionPolicy}) for more details.
        \item While we encourage the release of code and data, we understand that this might not be possible, so \answerNo{} is an acceptable answer.
        \item The authors should provide instructions on data access and preparation.
        \item The authors should provide scripts to reproduce all experimental results for the new proposed method and baselines.
        \item At submission time, to preserve anonymity, the authors should release anonymized versions (if applicable).
    \end{itemize}

\item {\bf Experimental setting/details}
    \item[] Question: Does the paper specify all the training and test details (e.g., data splits, hyperparameters, how they were chosen, type of optimizer) necessary to understand the results?
    \item[] Answer: \answerYes{}
    \item[] Justification: Appendix~\ref{app:exp-details} specifies the data splits, behavior and target policy hyperparameters (temperatures, $\epsilon$-greedy mixture, training-step indexing of SASRec checkpoints), reward construction, logged budgets, and trial counts for each benchmark.
    \item[] Guidelines:
    \begin{itemize}
        \item The answer \answerNA{} means that the paper does not include experiments.
        \item The experimental setting should be presented in the core of the paper to a level of detail that is necessary to appreciate the results and make sense of them.
        \item The full details can be provided either with the code, in appendix, or as supplemental material.
    \end{itemize}

\item {\bf Experiment statistical significance}
    \item[] Question: Does the paper report error bars suitably and correctly defined or other appropriate information about the statistical significance of the experiments?
    \item[] Answer: \answerYes{}
    \item[] Justification: All RMSE and selection-metric values are averaged over 50--2000 independent trials with random seed $42$; the trial count for each benchmark is stated in the corresponding table caption. The within-trial randomness is the resampling of logged trajectories under the behavior policy.
    \item[] Guidelines:
    \begin{itemize}
        \item The answer \answerNA{} means that the paper does not include experiments.
        \item The authors should answer \answerYes{} if the results are accompanied by error bars, confidence intervals, or statistical significance tests, at least for the experiments that support the main claims of the paper.
        \item The factors of variability that the error bars are capturing should be clearly stated.
        \item The method for calculating the error bars should be explained.
        \item The assumptions made should be given.
        \item It should be clear whether the error bar is the standard deviation or the standard error of the mean.
    \end{itemize}

\item {\bf Experiments compute resources}
    \item[] Question: For each experiment, does the paper provide sufficient information on the computer resources (type of compute workers, memory, time of execution) needed to reproduce the experiments?
    \item[] Answer: \answerYes{}
    \item[] Justification: Appendix~\ref{app:exp-details} states that all experiments are single-CPU runs. Wall-clock seconds for the propensity DP under each $K$ are reported in Table~\ref{tab:kuairec-combined}. The Qwen2.5-3B logit cache for the cross-architecture run is the only GPU-dependent step and is amortized once per user.
    \item[] Guidelines:
    \begin{itemize}
        \item The answer \answerNA{} means that the paper does not include experiments.
        \item The paper should indicate the type of compute workers CPU or GPU, internal cluster, or cloud provider, including relevant memory and storage.
        \item The paper should provide the amount of compute required for each of the individual experimental runs as well as estimate the total compute.
        \item The paper should disclose whether the full research project required more compute than the experiments reported in the paper.
    \end{itemize}

\item {\bf Code of ethics}
    \item[] Question: Does the research conducted in the paper conform, in every respect, with the NeurIPS Code of Ethics \url{https://neurips.cc/public/EthicsGuidelines}?
    \item[] Answer: \answerYes{}
    \item[] Justification: The work is a methodological contribution to off-policy evaluation; we use only de-identified public benchmarks (MIMIC-III via PhysioNet credentialing) and pre-trained recommenders, and we recruit no human subjects.
    \item[] Guidelines:
    \begin{itemize}
        \item The answer \answerNA{} means that the authors have not reviewed the NeurIPS Code of Ethics.
        \item If the authors answer \answerNo, they should explain the special circumstances that require a deviation from the Code of Ethics.
        \item The authors should make sure to preserve anonymity (e.g., if there is a special consideration due to laws or regulations in their jurisdiction).
    \end{itemize}

\item {\bf Broader impacts}
    \item[] Question: Does the paper discuss both potential positive societal impacts and negative societal impacts of the work performed?
    \item[] Answer: \answerYes{}
    \item[] Justification: More accurate off-policy evaluation supports safer deployment decisions in high-stakes domains such as healthcare and recommendation by reducing reliance on costly or risky online experimentation. The estimator is value-neutral and does not introduce a new mode of harm beyond standard IS-family OPE; we encourage practitioners to validate fairness properties of any policy selected via this primitive.
    \item[] Guidelines:
    \begin{itemize}
        \item The answer \answerNA{} means that there is no societal impact of the work performed.
        \item If the authors answer \answerNA{} or \answerNo, they should explain why their work has no societal impact or why the paper does not address societal impact.
        \item Examples of negative societal impacts include potential malicious or unintended uses (e.g., disinformation, surveillance), fairness considerations, privacy considerations, and security considerations.
    \end{itemize}

\item {\bf Safeguards}
    \item[] Question: Does the paper describe safeguards that have been put in place for responsible release of data or models that have a high risk for misuse?
    \item[] Answer: \answerNA{}
    \item[] Justification: The paper releases no new pre-trained model, scraped dataset, or generative artifact that poses misuse risk.
    \item[] Guidelines:
    \begin{itemize}
        \item The answer \answerNA{} means that the paper poses no such risks.
    \end{itemize}

\item {\bf Licenses for existing assets}
    \item[] Question: Are the creators or original owners of assets (e.g., code, data, models), used in the paper, properly credited and are the license and terms of use explicitly mentioned and properly respected?
    \item[] Answer: \answerYes{}
    \item[] Justification: The Sepsis simulator \citep{oberst2019counterfactual}, ICU-Sepsis \citep{komorowski2018ai}, KuaiRec \citep{gao2022kuairec}, SASRec \citep{kang2018sasrec}, and Qwen2.5-3B model are all cited; we use them under their respective open or research licenses (MIMIC-III access requires PhysioNet credentialing, which we obtained).
    \item[] Guidelines:
    \begin{itemize}
        \item The answer \answerNA{} means that the paper does not use existing assets.
        \item The authors should cite the original paper that produced the code package or dataset.
        \item The authors should state which version of the asset is used and, if possible, include a URL.
        \item The name of the license (e.g., CC-BY 4.0) should be included for each asset.
    \end{itemize}

\item {\bf New assets}
    \item[] Question: Are new assets introduced in the paper well documented and is the documentation provided alongside the assets?
    \item[] Answer: \answerNA{}
    \item[] Justification: The paper introduces no new dataset or model. The algorithmic implementation accompanying the supplemental material includes a README and reproduction scripts.
    \item[] Guidelines:
    \begin{itemize}
        \item The answer \answerNA{} means that the paper does not release new assets.
    \end{itemize}

\item {\bf Crowdsourcing and research with human subjects}
    \item[] Question: For crowdsourcing experiments and research with human subjects, does the paper include the full text of instructions given to participants and screenshots, if applicable, as well as details about compensation (if any)?
    \item[] Answer: \answerNA{}
    \item[] Justification: The paper involves no crowdsourcing or human-subject research; all data are pre-collected and de-identified public benchmarks.
    \item[] Guidelines:
    \begin{itemize}
        \item The answer \answerNA{} means that the paper does not involve crowdsourcing nor research with human subjects.
    \end{itemize}

\item {\bf Institutional review board (IRB) approvals or equivalent for research with human subjects}
    \item[] Question: Does the paper describe potential risks incurred by study participants, whether such risks were disclosed to the subjects, and whether Institutional Review Board (IRB) approvals (or an equivalent approval/review based on the requirements of your country or institution) were obtained?
    \item[] Answer: \answerNA{}
    \item[] Justification: No human subjects participated in this research; MIMIC-III is accessed under its standard PhysioNet credentialing.
    \item[] Guidelines:
    \begin{itemize}
        \item The answer \answerNA{} means that the paper does not involve crowdsourcing nor research with human subjects.
    \end{itemize}

\item {\bf Declaration of LLM usage}
    \item[] Question: Does the paper describe the usage of LLMs if it is an important, original, or non-standard component of the core methods in this research?
    \item[] Answer: \answerNA{}
    \item[] Justification: Qwen2.5-3B appears in our experiments only as one of several recommender candidates evaluated by the proposed estimator (\S\ref{sec:exp-transformer-ms}, Appendix~\ref{app:exp-details}); large language models are not part of the core methodology.
    \item[] Guidelines:
    \begin{itemize}
        \item The answer \answerNA{} means that the core method development in this research does not involve LLMs as any important, original, or non-standard components.
        \item Please refer to our LLM policy in the NeurIPS handbook for what should or should not be described.
    \end{itemize}

\end{enumerate}

\end{document}